\documentclass[10pt,twocolumn,letterpaper]{article}
\usepackage[pagenumbers]{cvpr}
\usepackage{graphicx,amsmath,amssymb,booktabs,multirow}
\usepackage{microtype}
\definecolor{cvprblue}{rgb}{0.21,0.49,0.74}
\usepackage[pagebackref,breaklinks,colorlinks,allcolors=cvprblue]{hyperref}
\usepackage[capitalize]{cleveref}
\def\confName{CVPR}
\def\confYear{2027}
\newcommand{\method}{VECD}
\newcommand{\suppapp}[1]{Appendix~#1 of the supplementary material}
\newcommand{\norm}[1]{\left\lVert #1\right\rVert}
\title{Show Me the Change You Want:\\Visual Example-Guided Change Detection}
\author{%
Yuan Qian\textsuperscript{1}\qquad
Jiaqi Cui\textsuperscript{2}\qquad
Jie Ma\textsuperscript{1}\thanks{This work was supported in part by the Natural Science Foundation of China under Grant 62101052; and in part by the Fundamental Research Funds for the Central Universities under Grant 2024JJ040. Corresponding author: Jie Ma (e-mail: \texttt{majie\_sist@bfsu.edu.cn}).}\\[4pt]
\parbox{0.96\textwidth}{\centering\small
\textsuperscript{1}School of Information Science and Technology, Beijing Foreign Studies University, Beijing 100089, China\\[2pt]
\textsuperscript{2}School of Mathematics, Statistics and Mechanics, Beijing University of Technology, Beijing, China}
}
\makeatletter
\apptocmd{\@maketitle}{%
  \begin{center}
  \captionsetup{hypcap=false}
  \includegraphics[width=0.85\textwidth]{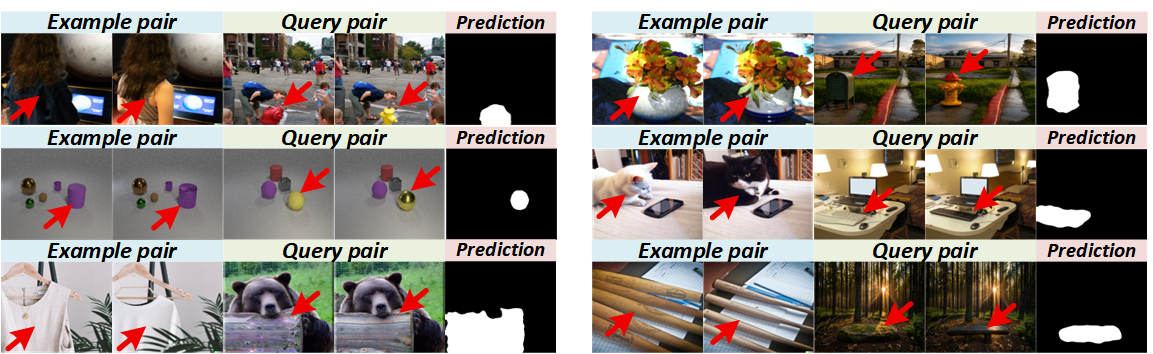}
  \captionof{figure}{\textbf{Show the change, not the object.} An unannotated example specifies an attribute change in another image pair. Red arrows highlight edits for the reader; they are not model inputs. White regions are predictions.}
  \label{fig:teaser}
  \end{center}
}{}{\PackageError{paper}{Cannot insert teaser}{Check the CVPR title macro.}}
\makeatother
\usepackage{tabularx,colortbl}
\definecolor{oursgray}{gray}{0.93}
\newcommand{\statcell}[2]{\mbox{\ensuremath{#1#2}}}
\begin{document}
\addtocontents{toc}{\protect\setcounter{tocdepth}{-1}}
\maketitle
\begin{abstract}
 Change detection identifies where images differ, but in practice users often seek only a particular kind of change. Existing referring change detection methods rely on predefined categories, textual descriptions, or annotated masks, limiting flexibility or requiring additional annotation. In this paper, we propose a new way to perform change detection, where an unannotated before-and-after image pair serves as a visual example to guide change localization in another pair. Our framework, Visual Example-Guided Change Detection (VECD), learns comparable attribute states from single images and introduces an example-conditioned readout to match changes across different objects and scenes. The example itself defines the request, without text, example masks, or test-time fine-tuning. We evaluate VECD on four datasets spanning rendered scenes and natural image edits: EditCLEVR, MagicBrush, HumanEdit, and MaskEdit. Experiments demonstrate cross-dataset generalization without fine-tuning: a single fixed example pair guides change detection across multiple scenes, even when examples and queries originate from different datasets. VECD selects matching changes in scenes with multiple simultaneous changes and returns an empty mask when the requested change is absent. It further generalizes to attribute families withheld from paired training, achieving 74.54\% to 93.38\% mIoU across four holdout settings.
Code is available at \url{https://github.com/qianyuancs/VECD}.
\end{abstract}

\section{Introduction}
\begin{figure}[t]
\centering\includegraphics[width=\columnwidth]{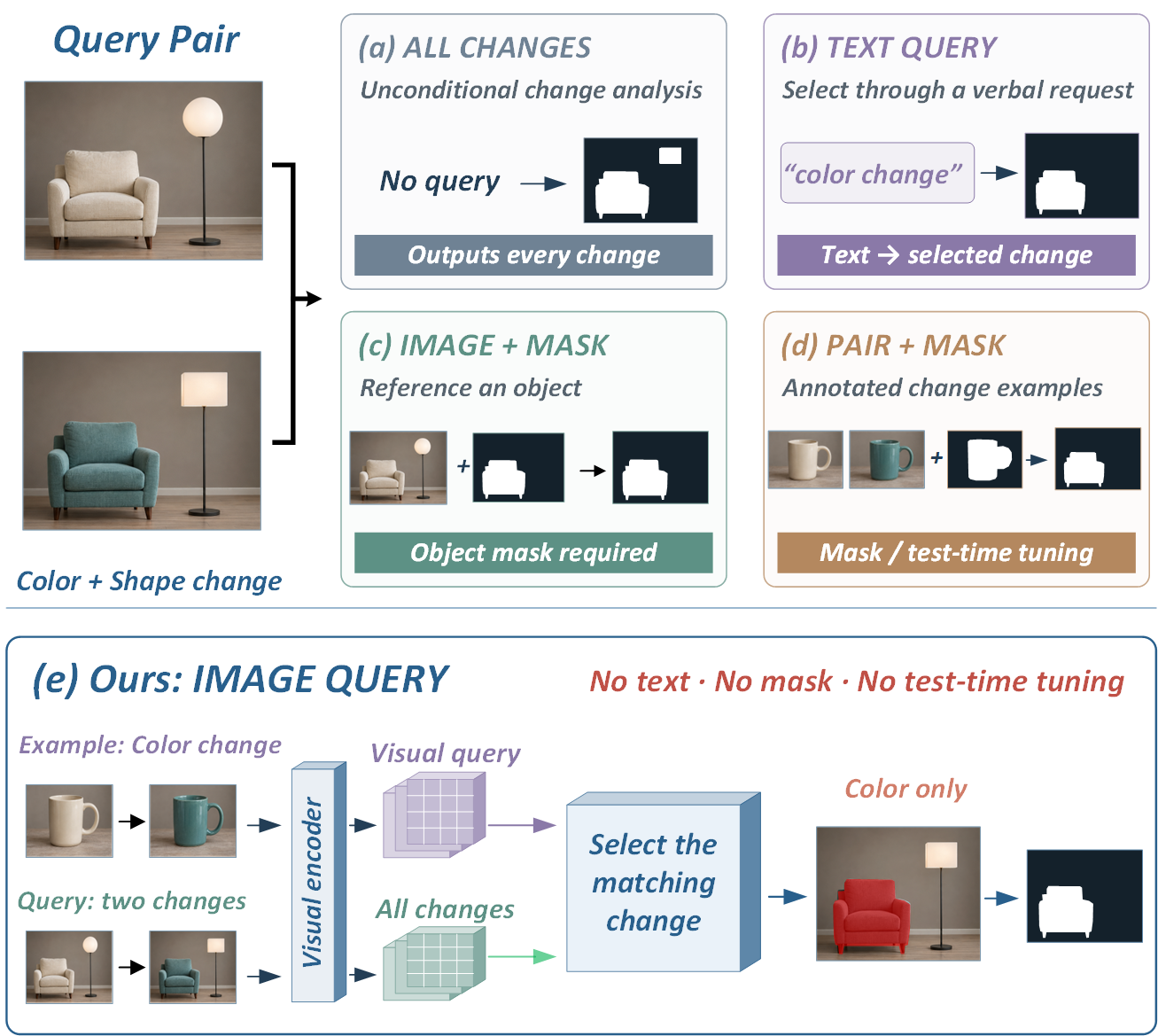}
\caption{\textbf{Methods for specifying the desired change.} (a) All-change detection has no relevance cue. (b) Text names the change. (c) A masked image specifies an object. (d) A labeled pair specifies a change. (e) VECD uses an unannotated pair. ``No query'' in (a) means no relevance request, not the absence of input images.}
\label{fig:interfaces}
\end{figure}

Change detection is a fundamental task for monitoring scenes, inspecting objects, and comparing their appearance~\cite{radke2005,sakurada2013,alcantarilla2016,changesim2021}. Conventional methods identify all changes between images~\cite{sakurada2015,gescf2025}. Users, however, often seek only one kind of change. \emph{For example, in \cref{fig:interfaces}, the chair changes from beige to teal, while the lampshade changes from round to rectangular.} A user looking for \textbf{color changes} would want only the chair highlighted. To meet this requirement, the detector must distinguish the requested change from other scene changes.

Referring change detection aims to localize only those changes that match the user's request~\cite{rcd2026}. There are two common ways to specify the desired change: (1) through a text instruction or (2) through both a visual example and a mask that marks the relevant region in the example.

Text-conditioned methods such as ViewDelta~\cite{viewdelta2025} allow requests beyond predefined semantic categories~\cite{sakurada2018}. Yet users must express visual differences in words. Broad changes are easy to describe, but precise descriptions of changes can be ambiguous or cumbersome. Recent work has therefore explored a visual alternative: an example image pair shows the change, and an accompanying mask marks the relevant region~\cite{filtering2023}.

Image-conditioned methods require the model to identify the change shown by the example pair. Related work on example-guided segmentation uses a mask to identify the region of interest. Matcher, SegIC, and DC-SAM~\cite{matcher2024,segic2024,dcsam2026}, for example, take an example image and its mask to find corresponding objects or regions in another image. The mask tells the model where to look in the example. These masks often require time-consuming manual annotation. Without an example mask, the detector faces a different challenge: it must infer which attribute changed from the two example images alone, then find the same kind of change in the query, even on a different object.

Another difficulty is recognizing the requested change across image domains. For instance, a rendered pair showing a color change could serve as an example for finding color changes in natural photographs. Rendered and natural images differ substantially in texture, lighting, and background complexity. The detector must therefore recognize the requested change despite these differences.

A further bottleneck is transferring a trained detector to a new dataset. One approach fine-tunes a detector on labeled examples of the requested change~\cite{filtering2023}. Such adaptation requires additional time and computational resources. Our goal is to use the same detector across datasets without updating its weights.

To address these issues, we propose \textbf{visual example-guided change detection}: an unannotated before-and-after pair specifies the attribute change to locate in another pair. Our framework, \method, transfers directly to other datasets without text prompts, example masks, or target-domain fine-tuning (\cref{fig:teaser,fig:interfaces}). Examples and queries can contain different objects and come from different image domains.

Specifically, \method\ first learns attribute representations, then learns to identify relevant changes from image pairs (\cref{fig:architecture}). An attribute-state reader learns static properties from individual images using frozen CLIP and DINOv2 features. An example-conditioned readout uses the example's change evidence to select query evidence. A mask decoder combines these responses and outputs the matching change mask. Our contributions are threefold:

\begin{itemize}
\item We propose \textbf{VECD}, a new framework for visual example-guided change detection. One unannotated example pair specifies the attribute change to locate among simultaneous changes, without any text prompts or example masks.

\item We propose an attribute-state reader that uses learnable attribute tokens to encode distinct properties at each location in a single image. We then construct an example-conditioned readout that uses the attribute changes in the example pair to select matching changes in the query pair. The model matches changes by the property that changed rather than by the identity of the object.

\item We demonstrate direct cross-dataset transfer without target-domain fine-tuning or test-time optimization. A single fixed example can guide detection across many scenes, even when the example and query come from different datasets and show different objects; in a scene with several changes, switching only the example lets VECD select the matching change. VECD also generalizes to attribute families excluded from paired training, achieving 74.54\%--93.38\% mIoU across four EditCLEVR holdout settings.
\end{itemize}

\section{Related work}
\label{sec:related}
\subsection{General change detection}
Change detection locates changes between observations of the same scene while suppressing illumination and viewpoint variation~\cite{radke2005}. Conventional detectors take no user query and return all task-defined changes. Their comparison strategies follow three broad directions. \emph{Explicit comparison} uses background modeling, geometric alignment, or feature differences to distinguish scene changes from camera and appearance variation~\cite{stauffer1999,sakurada2013,sakurada2015}. \emph{Supervised paired learning} trains Siamese or attention-based networks on annotated pairs to predict changed regions~\cite{bitcd2022,changeformer2022} and establish object correspondence~\cite{alcantarilla2016,daudt2018,cyws2023}. \emph{Pretrained-feature comparison} combines frozen representations with registration for cross-scene transfer~\cite{cyws3d2023}, or foundation-model features with SAM masks for training-free localization~\cite{gescf2025,sam2023}. Beyond localization, semantic prediction~\cite{sakurada2018} and change captioning~\cite{spotdiff2018,park2019,vacc2021,clip4idc2022} describe what changed through class labels or language. These richer outputs still do not select changes according to a user's example. Instead, VECD uses an unannotated example pair to specify the requested attribute change, locate the matching query regions, and suppress unrelated changes.

\subsection{Conditional change detection}
Conditional change detection adds an explicit relevance condition to image comparison: the output should contain only the specific changes requested by the user. Existing approaches specify this condition through either natural language prompts or annotated visual examples.

\emph{Text-conditioned methods} use verbal requests to select relevant changes. Unlike text-guided single-image segmentation~\cite{lseg2022,clipseg2022,catseg2024}, they compare two image states. Category-guided detectors use class-name embeddings in decoder cross-attention to select changes involving that class~\cite{rcd2026}. Free-form conditioning jointly processes text tokens, paired-image tokens, and learned segmentation queries in a transformer, then decodes the updated queries into a change mask~\cite{viewdelta2025}. This supports attribute and scene-level requests beyond category names. Both approaches still require users to describe the visual difference in words.

\emph{Annotated-example methods} specify the desired change through labeled before-and-after pairs. Few-shot specialization trains a general change detector, then fine-tunes it on a few mask-labeled examples~\cite{filtering2023}. Obtaining these masks is a practical bottleneck: ordinary image pairs do not include change annotations, so users must first delineate the relevant pixels. This turns specifying a change into an additional labeling task. VECD instead reads the change directly from an unannotated example pair. It requires neither text prompts nor example masks and processes new query pairs without test-time fine-tuning.

\subsection{Visual prompting}
Visual prompting turns image examples into instructions. The growing use of examples across visual tasks has motivated different mechanisms for extracting relevant information from examples to guide predictions on new images.

\emph{Images as content queries.} An example image and mask identify what to segment in another image~\cite{oslsm2017}. The masked features guide prediction through region prototypes~\cite{ppnet2020} and iterative comparison~\cite{panet2019,canet2019}, or spatial matching across images~\cite{pfenet2022,hsnet2021,cyctr2021}. Foundation models convert these correspondences into decoder prompts or visual instructions~\cite{matcher2024,segic2024,dcsam2026,persam2024}. In-context methods instead arrange example images and outputs in a shared format, from which the model predicts the query output~\cite{visualprompt2022,painter2023,seggpt2023}.

\emph{Image pairs as operation prompts.} Text-guided editors encode a written instruction~\cite{instructpix2023,emuedit2024}; example-based editors obtain an editing condition from a before-and-after pair. Methods fit an instruction embedding to the pair~\cite{visii2023} or transfer the demonstrated edit through a generator's latent space~\cite{editone2024}. Learned approaches read the pair as diffusion context~\cite{promptdiffusion2023,imagebrush2023}, summarize its operation in manipulation tokens~\cite{instamanip2025}, or learn a pair embedding through contrastive pretraining~\cite{editclip2025}. The visual condition then guides how to edit a new image.

Change detection guided by visual examples remains underexplored,
particularly when the same example must work across scenes and image
domains. VECD addresses this challenge by matching attribute changes
rather than object identities, so one example pair can guide detection
in new scenes and across datasets without fine-tuning.

\section{Method}
\label{sec:method}
Given an unannotated before-and-after example pair, VECD localizes changes of the same type in a query pair. For example, a mug turning blue can serve as an example for finding a car that turns green: the two pairs share a color change, but neither the object nor the colors need to match.

Let $E=(E^0,E^1)$ and $Q=(Q^0,Q^1)$ denote the before-and-after example and query pairs. VECD predicts a mask for the query pair
\begin{equation}
 \hat{Y}_Q=f_\theta(E^0,E^1,Q^0,Q^1).
\end{equation}
The example specifies a change in color, size, shape, or material. Only query regions with that specific change should be detected; if none match, the output mask is empty. All four input images are unannotated RGB images.

\begin{figure*}[t]
 \centering
 \includegraphics[width=\textwidth]{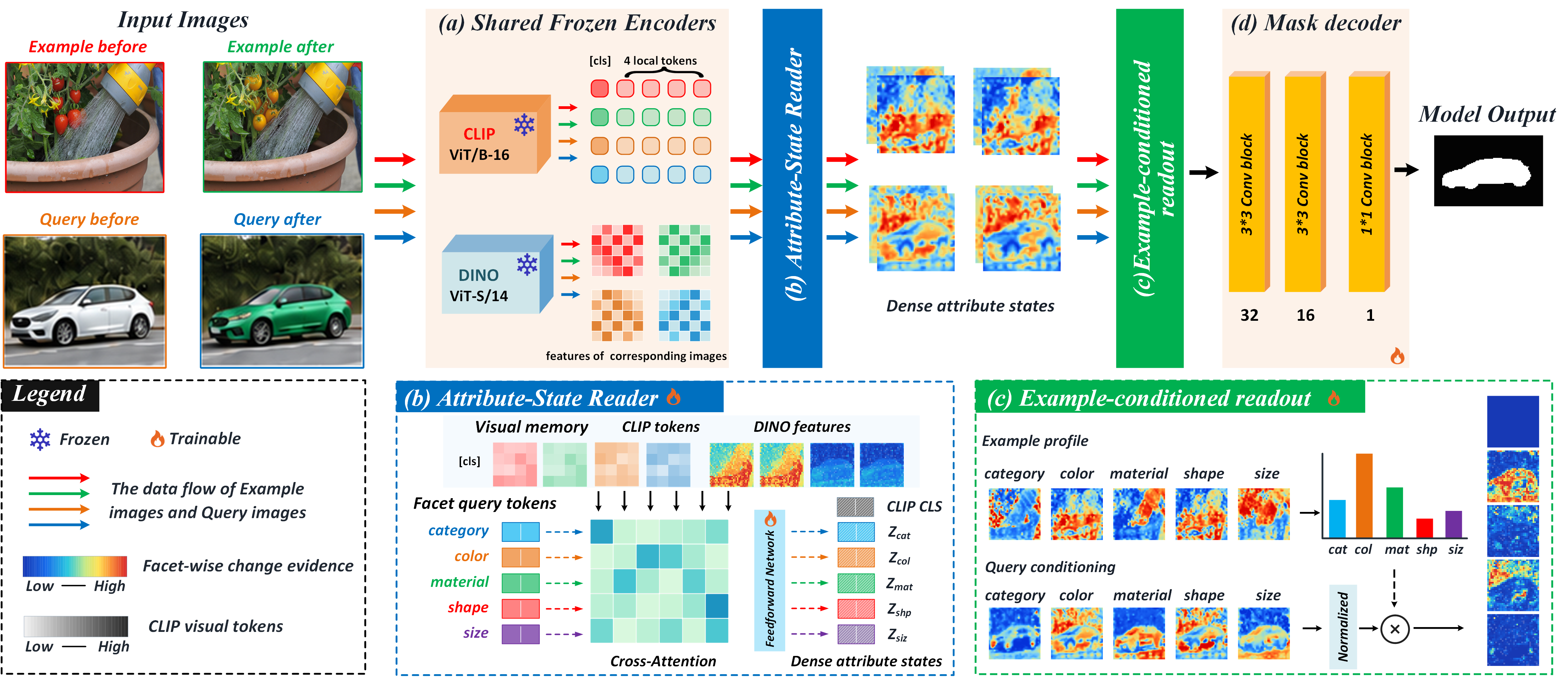}
 \caption{\textbf{VECD.} (a) All four images pass through the same frozen encoders. (b) At each position, five facet queries read the CLIP tokens of the enclosing tile and the DINOv2 token at that position. (c) The example yields one change profile for the whole image pair. (d) The mask decoder combines the matching map with spatial cues from the query pair to predict the requested change mask.}
 \label{fig:architecture}
\end{figure*}

\cref{fig:architecture} shows the overall framework. Shared encoders and an attribute-state reader link image features to specific attributes. The readout summarizes the demonstrated change across attributes and selects matching change evidence in the query pair. The mask decoder helps generate a corresponding binary change mask.

\subsection{Shared frozen encoders}
We encode all four images using the same two frozen, pretrained vision transformers~\cite{vit2021}: CLIP~\cite{clip2021} and DINOv2~\cite{dinov2}. CLIP provides appearance context for each tile, while DINOv2 preserves local detail on a spatial grid~\cite{dino2021,maskclip2022}. First, CLIP ViT-B/16~\cite{clip2021} processes the $512\times512$ image in four equal tiles, each resized to $224\times224$. Each tile yields a CLS token and four local tokens, obtained by averaging each $7\times7$ quadrant of its patch grid before CLIP's final normalization and projection. DINOv2 ViT-S/14~\cite{dinov2} processes the whole image at $448\times448$.

For each image, CLIP produces a tensor $\mathbf{C} \in \mathbb{R}^{4 \times 5 \times 512}$. DINOv2 produces a tensor $\mathbf{D} \in \mathbb{R}^{32 \times 32 \times 384}$, with each position holding a 384-dimensional descriptor.

\subsection{Attribute-state reader}
The attribute-state reader uses learnable query tokens to extract local attribute states. As shown in \cref{fig:architecture}, at each spatial location $p$, the reader combines five CLIP tokens
$c_{p,1}, \ldots, c_{p,5}$ and one DINOv2 feature $d_p$ into a
six-token visual memory. We project the normalized features to 512
dimensions and concatenate them along the token dimension:
\begin{equation}
    M_p = \left[
        W_C c_{p,1}, \ldots, W_C c_{p,5}, W_D d_p
    \right] + P \in \mathbb{R}^{6 \times 512}.
    \label{eq:visual_memory}
\end{equation}
Here $W_C$ and $W_D$ are learnable projection matrices for CLIP and
DINOv2 features, respectively, and $P$ is a learned embedding for each token slot.

We design five learnable attribute tokens to represent category, color, material, shape, and size.
Each token queries the local visual memory through cross-attention to obtain an attribute-specific representation.
The five projected queries (see \suppapp{B}) form the rows of $U\in\mathbb{R}^{5\times512}$. At each position \(p\), we apply multihead cross-attention~\cite{attention2017} with \(U\) as the queries and \(M_p\) as both keys and values:
\begin{align}
H_p    &= U + \operatorname{MHA}(U, M_p, M_p), \\
H'_p   &= H_p + \operatorname{FFN}(\operatorname{LN}(H_p)), \\
Z_{p,k}&= \operatorname{norm}\!\left(
           c_{p,1} + W_O H'_{p,k} + b_O \right),
\end{align}
where MHA denotes multihead cross-attention, LN denotes layer normalization~\cite{layernorm2016}, and FFN is a two-layer feed-forward network with GELU activation~\cite{gelu2016}.
The reader produces an attribute-state tensor $Z\in\mathbb{R}^{32\times32\times5\times512}$ for each image. $k=1,\ldots,5$ indexes category, color, material, shape, and size, respectively.

\subsection{Example-conditioned readout}
Given the attribute states, the readout selects relevant changes in the query pair by matching local attribute changes to the pattern demonstrated by the example. Specifically, we first summarize how much each attribute changes in the example pair. We use this profile to weight attribute changes at each location in the query pair.

For each pair $X$, we measure the squared Euclidean distance between the normalized before and after states on the matching grid:
\begin{equation}
 \mathrm{Diff}^X_k(p)=\norm{z^{X,1}_{p,k}-z^{X,0}_{p,k}}_2^2,
 \qquad X\in\{E,Q\},
\end{equation}
where \(\mathrm{Diff}_{k}^{X}(p)\) denotes the magnitude of change in attribute \(k\) at location \(p\) between the two images in pair \(X\).

For the example pair, we average the change response of each attribute \(k\) over all \(N=32^2\) spatial locations. This gives a mean response \(s_k\) for each attribute. Normalizing these responses across attributes yields \(\alpha_k\), the relative weight of attribute \(k\) in the demonstrated change:
\begin{equation}
 s_k=\frac{1}{N}\sum_p \mathrm{Diff}^E_k(p),\qquad
 \alpha_k=\frac{s_k}{\max(\sum_j s_j,\epsilon)} .
\end{equation}
As for the query images, we normalize the change responses across attributes at each location \(p\), yielding a local weight \(\beta_k(p)\) for each attribute \(k\). Both denominators sum over facets $j$ and use $\epsilon=10^{-8}$:
\begin{equation}
 \beta_k(p)=\frac{\mathrm{Diff}^Q_k(p)}{\max(\sum_j \mathrm{Diff}^Q_j(p),\epsilon)} .
 \label{eq:query-profile}
\end{equation}
The match score $r_k(p)=\alpha_k\beta_k(p)$ is high when attribute $k$ accounts for a large share of the change in both pairs.

To preserve local change magnitude, the readout incorporates the unnormalized query response \(\mathrm{Diff}_k^Q(p)\). We also use the geometric mean
$v_k(p)=\sqrt{s_k\,\mathrm{Diff}^Q_k(p)}$ to capture the strength of the change in both the example and the query. We combine all these features into a single input for the shared facet readout:
\begin{equation}
 x_k(p)=[\mathrm{Diff}^Q_k(p),\alpha_k,\beta_k(p),r_k(p),v_k(p)] .
 \label{eq:facet-input}
\end{equation}
For each attribute $k$, the shared readout $\phi$ applies two $3\times3$
convolutions with GroupNorm~\cite{groupnorm2018} and GELU, followed by
a $1\times1$ projection, to produce a feature map $F_k=\phi(x_k)$.
At each query location $p$, we concatenate the match-weighted sum and
the channel-wise maximum across attributes to form the matching map
$H_Q\in\mathbb{R}^{32\times32\times16}$:
\begin{equation}
\begin{aligned}
\bar r_k(p)
&= \frac{r_k(p)}{\max\!\left(\sum_j r_j(p),\,\epsilon\right)}, \\
H_Q(p)
&= \left[\sum_k \bar r_k(p)\,F_k(p)\,;\,\max_k F_k(p)\right].
\end{aligned}
\label{eq:match-map}
\end{equation}

\subsection{Mask decoder}
The mask decoder combines the matching map $H_Q$ with spatial cues from
the query pair. Let $\mathbf{D}_Q^0$ and $\mathbf{D}_Q^1$ denote the
DINOv2 feature grids of the before and after images, respectively.
From their mean and absolute difference, we compute
a spatial feature map and predict the probability of any change at
each location.

We concatenate the upsampled matching map, the spatial features, and the
change probabilities as input to a five-layer convolutional decoder.
It consists of two blocks with two $3\times3$ convolutions each,
followed by a $1\times1$ convolution that predicts a mask logit at
each location. We resize and threshold the resulting predictions to
obtain the binary mask $\hat{Y}_Q$ of regions that match the
demonstrated change.

\subsection{Training objectives}
We train VECD in two stages: the reader learns what each attribute
looks like in a single image, and the detector then learns where the
attribute change shown in an example occurs in a query pair.
\paragraph{Learning states from single images.}
The static loss trains the reader to identify attribute values. For example, a color label
specifies whether a region is red or blue. For each attribute $k$, we encode its possible values with
the frozen CLIP text encoder and stack the normalized embeddings as
rows of $B_k$. The product $Z_k B_k^\top$ then scores each possible
value at each location.

To preserve the reader's pretrained representation while learning attribute labels, we keep a frozen copy of the reader as a teacher.
Given the same image, $\mathcal L_{\mathrm{teacher}}$ measures the
mean cosine distance between the two readers' state vectors at matching locations and facets.
The static loss is
\begin{equation}
\mathcal L_{\mathrm{static}}
= \underset{k\in\mathcal F}{\operatorname{mean}}\,
  \operatorname{CE}\!\left(Z_k B_k^\top/T, y_k\right)
  + 0.25\,\mathcal L_{\mathrm{teacher}}.
\end{equation}

\paragraph{Learning to select changes from pairs.}
A query pair may contain several changes, but only those that match
the example should appear in the output. We therefore use
$\mathcal L_{\mathrm{query}}$ to supervise the final prediction with
the mask of the requested change, or an empty mask if that change
is absent. Two auxiliary losses guide the selection.
$\mathcal{L}_{\mathrm{select}}$ measures the difference between the selector's prediction and the mask of the demonstrated change in the example pair.
$\mathcal{L}_{\mathrm{change}}$ compares the predicted change map for each pair with a mask of all changed regions in that pair. We freeze the reader and train the remaining modules with
\begin{equation}
\mathcal L_{\mathrm{pair}}
= \mathcal L_{\mathrm{query}}
+ 0.5\,\mathcal L_{\mathrm{select}}
+ 0.25\,\mathcal L_{\mathrm{change}}.
\end{equation}
Each term combines binary cross-entropy with Dice loss~\cite{vnet2016}. \suppapp{C} illustrates the supervision masks.
\section{Experiments}
\label{sec:experiments}

\paragraph{Implementation details.}
We run our experiments on four NVIDIA RTX 4090 GPUs (24\,GB each). We compute and save the frozen CLIP and DINOv2 features before training. We use AdamW~\cite{adamw2019} with learning rate $10^{-4}$ and weight decay $10^{-3}$. Single-image training runs for two epochs with an effective batch size of four. Paired training runs for 20 epochs with batch size eight and equal numbers of EditCLEVR and MagicBrush requests. We report means over three mixed-training seeds for VECD and its ablations.
\begin{table*}[!t]
\caption{\textbf{Same-dataset conditional change detection (\%).} Examples are drawn anew for each query and attribute. Each prediction uses one example pair. $M$: example mask; $A$: example-only adaptation; $T$: text; $U$: unconditional. Ours uses no annotation. \textbf{Bold}/\underline{underline}: best/second-best per metric.}
\label{tab:main}\centering\footnotesize
\setlength{\tabcolsep}{1pt}
\begin{tabular*}{\textwidth}{@{\extracolsep{\fill}}lrrrrrrrrrrrrrrrr@{}}
\toprule
Method & \multicolumn{4}{c}{EditCLEVR} & \multicolumn{4}{c}{MagicBrush} & \multicolumn{4}{c}{HumanEdit} & \multicolumn{4}{c}{MaskEdit}\\
\cmidrule(lr){2-5} \cmidrule(lr){6-9} \cmidrule(lr){10-13} \cmidrule(lr){14-17}
 & mIoU & F1 & Re & Pre & mIoU & F1 & Re & Pre & mIoU & F1 & Re & Pre & mIoU & F1 & Re & Pre\\\midrule
Filtering$^{M,A}$ & 54.32 & 18.15 & 31.08 & 12.84 & 55.17 & 32.18 & 52.58 & 23.23 & 49.87 & 24.23 & 49.28 & 16.08 & 50.29 & 23.63 & \underline{37.97} & 17.23\\
Matcher$^M$ & 50.89 & 7.98 & 21.36 & 4.90 & 44.09 & 3.00 & 5.27 & 2.10 & 40.51 & 2.57 & 5.98 & 1.64 & 44.11 & 4.96 & 7.32 & 3.75\\
SegIC$^M$ & 58.68 & 31.68 & \underline{69.85} & 20.51 & 52.44 & 23.57 & 33.18 & 18.64 & 52.19 & 24.62 & 34.98 & 19.12 & 50.10 & 18.02 & 20.98 & 15.98\\
DC-SAM$^M$ & 50.94 & 9.27 & 31.76 & 5.43 & 46.84 & 6.41 & 8.41 & 5.18 & 47.79 & 11.04 & 14.18 & 9.03 & 46.72 & 12.11 & 17.45 & 9.27\\
\midrule
ViewDelta$^T$ & \underline{60.17} & \underline{35.29} & 62.91 & \underline{24.75} & \underline{55.64} & 28.27 & 25.90 & \textbf{31.46} & \underline{54.02} & 25.80 & 26.24 & \textbf{25.67} & 52.48 & 18.94 & 12.65 & \textbf{38.65}\\
GeSCF$^U$ & 57.97 & 29.03 & 44.37 & 21.57 & 54.23 & \underline{32.85} & \underline{68.07} & 21.65 & 53.38 & \underline{32.03} & \underline{63.28} & 21.44 & \underline{54.75} & \underline{27.09} & 22.24 & \underline{34.65}\\
\midrule
Ours & \textbf{91.01} & \textbf{90.17} & \textbf{91.05} & \textbf{89.38} & \textbf{56.03} & \textbf{37.22} & \textbf{79.29} & \underline{24.32} & \textbf{54.55} & \textbf{36.20} & \textbf{79.79} & \underline{23.41} & \textbf{58.09} & \textbf{38.12} & \textbf{43.55} & 33.92\\
\bottomrule
\end{tabular*}
\end{table*}
\subsection{Data and evaluation protocols}
\paragraph{Training and test sets.}
We train on 9,808 EditCLEVR pairs~\cite{editclevr2026}, 128 additional pairs with multiple changes, and 97 MagicBrush pairs~\cite{magicbrush2023}. Single-image training uses 9,642 EditCLEVR frames. The test sets (examples in \suppapp{A}) contain 2,997 EditCLEVR, 127 MagicBrush, 454 HumanEdit~\cite{humanedit2024}, and 358 MaskEdit~\cite{maskflow2026} pairs. For the three natural-image datasets, we select pairs whose edit captions describe a change in color, size, shape, or material. For each such attribute, the edit-region mask serves as proxy ground truth. A separate set has 60 EditCLEVR pairs with multiple changes. Training, validation, calibration, and test sets share no source images or scenes.

\paragraph{Example sampling.}
Each prediction uses \textbf{one example pair and one query pair}. All visual methods share the sampled pairs; self-pairs and images from the same scene or editing chain are excluded.

\paragraph{Metrics.}
We report binary mIoU, F1, recall, and precision. Binary mIoU is the mean of foreground and background IoU. Metrics are computed from pooled pixel counts over all requests, including empty targets, and averaged across runs.

\subsection{Comparison methods}
We compare methods that use visual examples, text, or no query condition (\cref{tab:main}). Trainable baselines use the same training data and their own training procedures.

Filtering~\cite{filtering2023} fine-tunes a separate copy of its trained detector on each example pair and its mask. Matcher~\cite{matcher2024} matches the before images and the after images separately, then keeps pixels selected in both predictions. For SegIC~\cite{segic2024} and DC-SAM~\cite{dcsam2026}, we place each image pair side by side and copy the example mask to both frames. We align and average their two query-frame logits. The segmentation networks are unchanged; Matcher needs no training.

ViewDelta~\cite{viewdelta2025} receives explicit text naming the requested attribute. GeSCF~\cite{gescf2025} receives only the raw query pair, so its prediction is the same for every request. All evaluated methods use the same targets, including empty masks. Only Filtering updates its weights at test time, using the example and its mask, not query labels.

\begin{figure}[t]
\centering\includegraphics[width=\columnwidth]{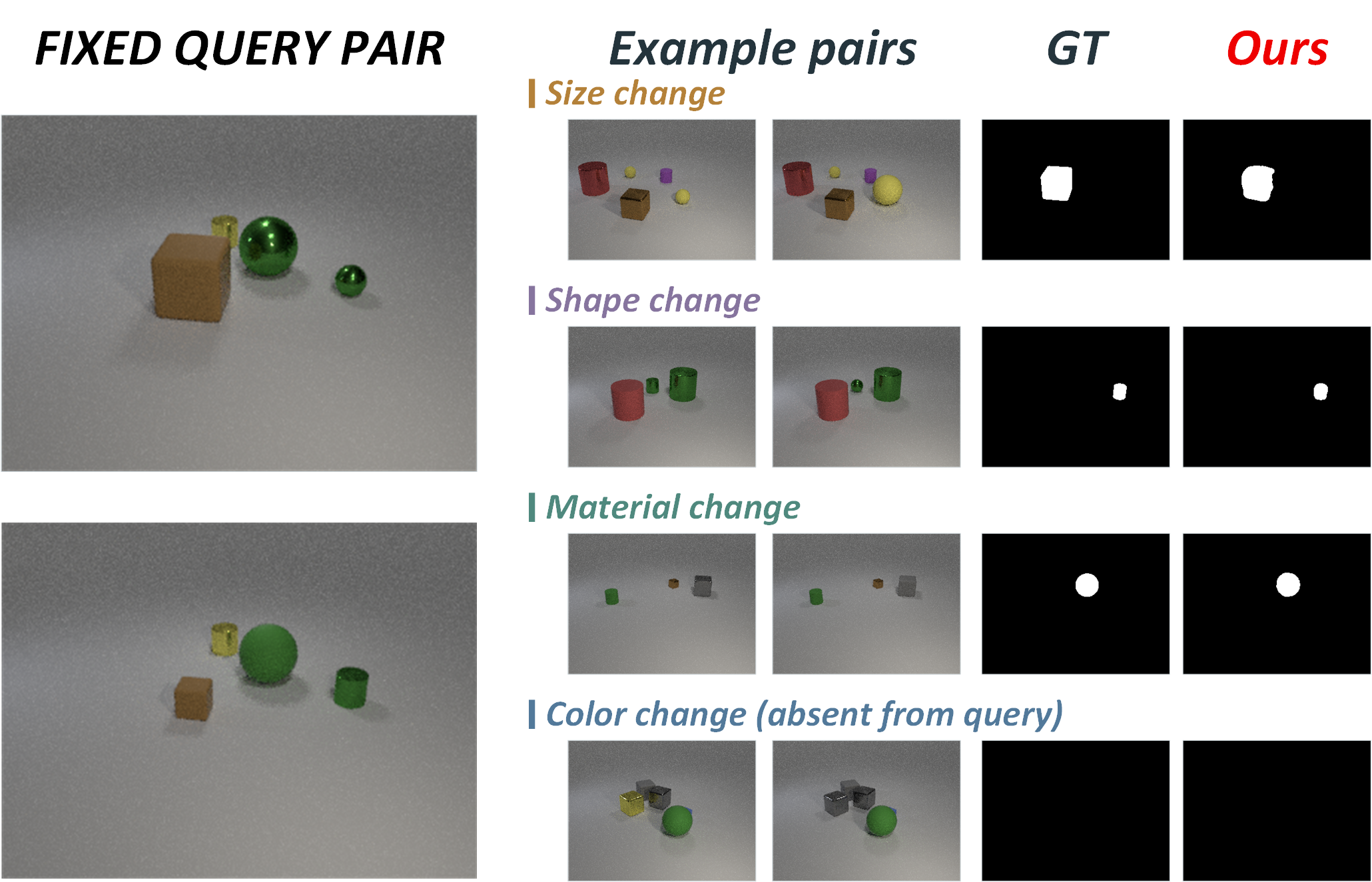}
\caption{\textbf{Same query, different examples.} Each row uses the query pair on the left and one example pair. The color example yields an empty mask because the query contains no color change.}
\label{fig:switch}
\end{figure}

\subsection{Conditional change detection}

\paragraph{Queries with one change.}
To test whether VECD can locate a requested change and reject nonmatching requests, we use EditCLEVR examples to query 2,997 EditCLEVR test pairs, each with one attribute change. For each query, we draw four examples, one per attribute, and test them one at a time: 11,988 requests per run. The matching example should return the changed object's mask; the other three should return empty masks. VECD reaches 91.01\% mIoU and 90.17\% F1, outperforming all six baselines (\cref{tab:main}). \suppapp{D} reports the standard deviations of mIoU and F1 across three training seeds.

\paragraph{Same-dataset natural queries.}
We repeat this same-dataset protocol on natural images, keeping weights and thresholds fixed. VECD reaches 56.03\%, 54.55\%, and 58.09\% mIoU on MagicBrush, HumanEdit, and MaskEdit, leading in mIoU and F1 without example masks (\cref{tab:main}).

\begin{table}[t]
\caption{\textbf{Multi-change selection.} FG IoU: foreground IoU; D-FPR: false-positive rate on other changed pixels ($\downarrow$). Notation follows \cref{tab:main}.}
\label{tab:native}\centering\small\setlength{\tabcolsep}{1.8pt}
\begin{tabular*}{\columnwidth}{@{\extracolsep{\fill}}lrrrr@{}}
\toprule
Method & mIoU & FG IoU & F1 & D-FPR$\downarrow$\\\midrule
Filtering$^{M,A}$ & 54.68 & 12.36 & 21.99 & 27.71\\
Matcher$^M$ & 53.28 & 9.19 & 16.83 & 12.34\\
SegIC$^M$ & \underline{55.94} & \underline{14.54} & \underline{25.39} & 23.65\\
DC-SAM$^M$ & 53.75 & 10.91 & 19.68 & 24.23\\
ViewDelta$^T$ & 52.42 & 6.74 & 11.90 & \underline{4.82}\\
\midrule
Ours & \textbf{92.20} & \textbf{84.71} & \textbf{91.71} & \textbf{1.74}\\
\bottomrule
\end{tabular*}
\end{table}

\paragraph{Selecting among simultaneous changes.}
To test whether VECD uses the example to select among multiple changes, we keep each query pair fixed and vary only the example. We test each of 60 multi-change EditCLEVR pairs with four fixed examples, one per attribute (240 requests per run). As shown in \cref{fig:switch}, each example selects its matching change while excluding other changes in the same scene.

VECD reaches 92.20\% mIoU (sample standard deviation: 0.89 points), compared with SegIC's 55.94\% (\cref{tab:native}). Its false-positive rate on other changed pixels is 1.74\%, versus 23.65\% for SegIC. Changing only the example lets VECD accurately select the requested change and reject the others in a scene with multiple simultaneous changes.

\subsection{Cross-dataset queries and visual transfer}
To test cross-domain generalization, we use MagicBrush or EditCLEVR examples to query HumanEdit and MaskEdit. For each query, we randomly sample one example per attribute from the selected source and test the four separately. Each source gives 1,816 HumanEdit and 1,432 MaskEdit requests per run. VECD keeps its weights and source-calibrated thresholds fixed, without target training. Filtering fine-tunes on each labeled example.

\begin{table}[!t]
\caption{\textbf{Cross-source visual queries (\%).} Models, query sets, and notation follow \cref{tab:main}.}
\label{tab:cross}\centering\small\setlength{\tabcolsep}{2pt}
\begin{tabular*}{\columnwidth}{@{\extracolsep{\fill}}lrrrr@{}}
\toprule
Method & \multicolumn{2}{c}{HumanEdit} & \multicolumn{2}{c}{MaskEdit}\\
\cmidrule(lr){2-3}\cmidrule(lr){4-5}
 & mIoU & F1 & mIoU & F1\\\midrule
\multicolumn{5}{c}{\textbf{MagicBrush examples}}\\\midrule
Filtering$^{M,A}$ & \underline{52.92} & \underline{28.73} & \underline{52.22} & \underline{27.30}\\
Matcher$^M$ & 40.58 & 3.20 & 43.52 & 2.62\\
SegIC$^M$ & 52.32 & 24.87 & 51.49 & 20.94\\
DC-SAM$^M$ & 46.89 & 10.02 & 48.33 & 14.79\\
\midrule
Ours & \textbf{54.63} & \textbf{36.39} & \textbf{58.03} & \textbf{37.97}\\
\midrule
\multicolumn{5}{c}{\textbf{EditCLEVR examples}}\\\midrule
Filtering$^{M,A}$ & 47.37 & 12.94 & 48.41 & 16.43\\
Matcher$^M$ & 46.23 & 1.62 & 47.37 & 3.91\\
SegIC$^M$ & \underline{52.43} & \underline{24.32} & \underline{51.32} & \underline{19.50}\\
DC-SAM$^M$ & 48.32 & 3.53 & 47.86 & 2.94\\
\midrule
Ours & \textbf{54.72} & \textbf{36.36} & \textbf{58.01} & \textbf{37.67}\\
\bottomrule
\end{tabular*}
\end{table}

\begin{figure}[!t]
\centering
\includegraphics[width=\columnwidth]{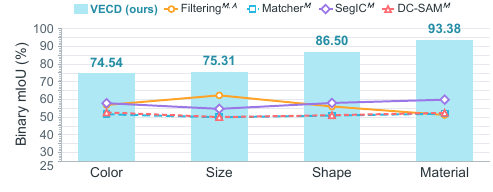}
\caption{\textbf{EditCLEVR family holdouts.}
Notation follows \cref{tab:main}.}
\label{fig:holdout}
\end{figure}

VECD leads all four visual baselines with both example sources (\cref{tab:cross}). MagicBrush examples yield 54.63\% and 58.03\% mIoU on HumanEdit and MaskEdit, with F1 gains of 7.66 and 10.67 points over Filtering. EditCLEVR examples perform similarly. VECD thus transfers attribute-change queries across domains, even from rendered examples to natural images, without example masks or weight updates. ViewDelta and GeSCF do not read examples and are omitted.

\begin{figure*}[t]
\centering
\includegraphics[width=\textwidth]{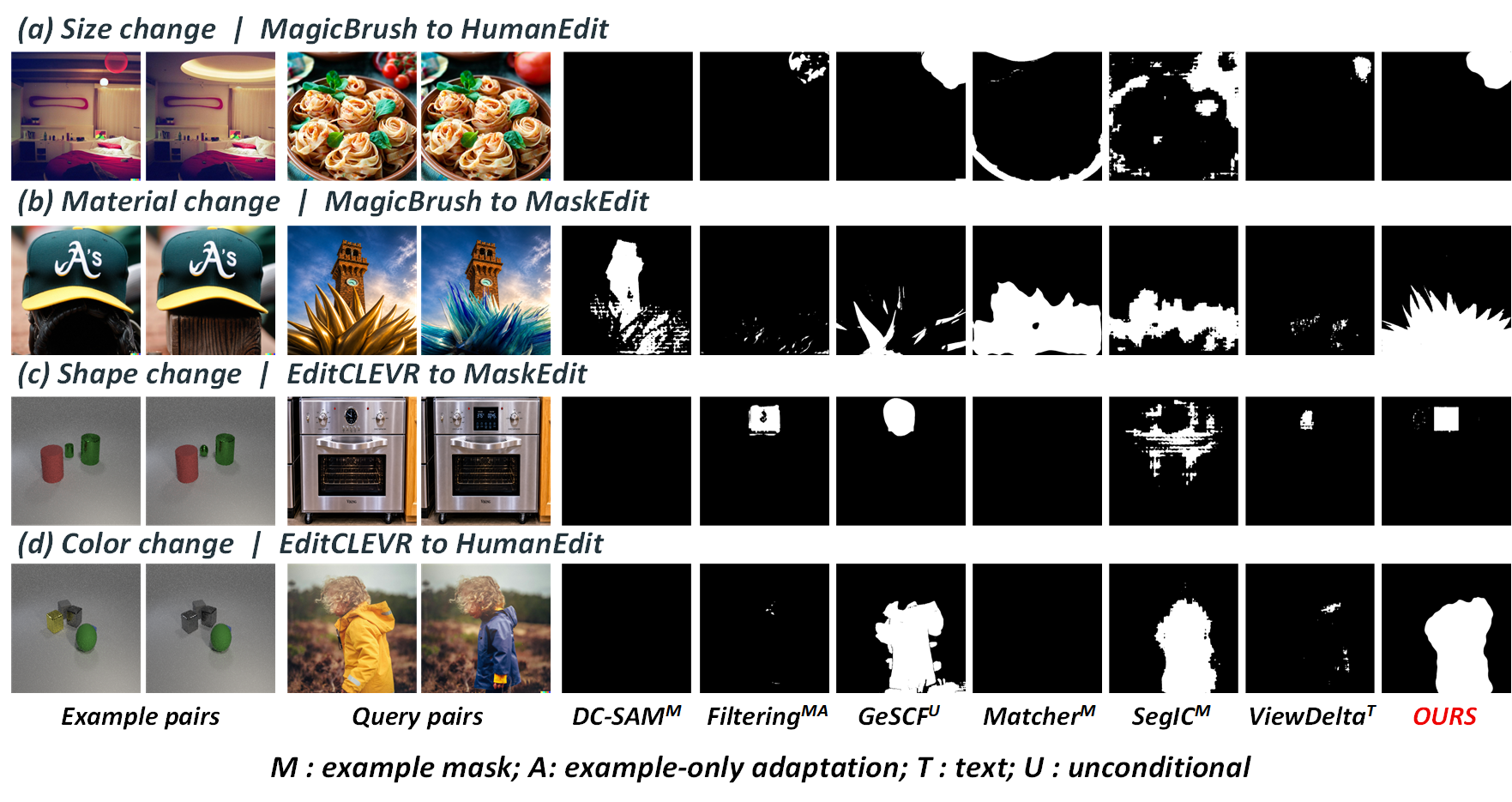}
\caption{\textbf{Visual querying across datasets.} MagicBrush and EditCLEVR examples query HumanEdit and MaskEdit.}
\label{fig:natural}
\end{figure*}

\subsection{Unseen paired change families}
\label{sec:holdout}
To test transfer to unseen change families, we train four EditCLEVR models, each with one family excluded from paired training. Pairs that change this attribute are removed from both positive and negative training requests. At test time, an example of the left-out family queries all 2,997 test pairs. Trainable visual baselines use the same splits; Filtering still fine-tunes on the labeled example.

VECD outperforms the strongest baseline by 12.99--33.51 mIoU points across the four holdout tests (\cref{fig:holdout}). These gains show strong generalization to change families unseen during paired training: VECD can use learned static states to detect changes without training pairs from those families.

\subsection{One example, many query images}
When users search for the same change across multiple images, they should be able to reuse a single example rather than select a new one for each query. We test this with eight held-out MagicBrush examples, two per attribute, randomly selected. Each separately queries all 454 HumanEdit and 358 MaskEdit pairs (6,496 requests per run, including empty targets). All methods use the same pairs. The baselines receive example masks, and Filtering also fine-tunes once per example. VECD needs neither. \suppapp{E} shows same-dataset queries and fixed-example predictions.

VECD leads all four visual baselines in mIoU and F1 (\cref{tab:fixed}), with F1 gains of 7.18 and 13.39 points over Filtering. These three-seed scores use fixed examples, unlike the random-example tests in \cref{tab:main,tab:cross}. A fixed example therefore remains effective across a full query set, so users need not choose a new example for each scene.

\begin{table}[!htbp]
\caption{\textbf{One example, many queries (\%).}}
\label{tab:fixed}\centering\small\setlength{\tabcolsep}{2pt}
\begin{tabular*}{\columnwidth}{@{\extracolsep{\fill}}lrrrr@{}}
\toprule
Method & \multicolumn{2}{c}{HumanEdit} & \multicolumn{2}{c}{MaskEdit}\\
\cmidrule(lr){2-3}\cmidrule(lr){4-5}
 & mIoU & F1 & mIoU & F1\\\midrule
Filtering$^{M,A}$ & \underline{54.32} & \underline{29.27} & \underline{52.38} & \underline{24.41}\\
Matcher$^M$ & 40.89 & 2.01 & 43.10 & 2.61\\
SegIC$^M$ & 51.75 & 24.76 & 51.04 & 21.05\\
DC-SAM$^M$ & 45.59 & 8.00 & 46.63 & 12.02\\
\midrule
VECD (ours) & \textbf{54.66} & \textbf{36.45} & \textbf{57.98} & \textbf{37.80}\\
\bottomrule
\end{tabular*}
\end{table}

\subsection{Ablation study}
Our ablation study tests which design choices help VECD locate the change specified by the example. We vary example pooling, visual cues, and supervision while keeping the training budget and test examples fixed (\cref{tab:ablation}; mIoU standard deviations across three training seeds in \suppapp{F}).
\begin{table}[!htbp]
\caption{\textbf{Ablation study.} Binary mIoU (\%), averaged over three training seeds. Below the divider, each variant changes one component of the learned-pooling model.}
\label{tab:ablation}\centering\footnotesize\setlength{\tabcolsep}{1.5pt}
\begin{tabular*}{\columnwidth}{@{\extracolsep{\fill}}lrrrr@{}}
\toprule
Variant & EditCLEVR & MagicBrush & HumanEdit & MaskEdit\\\midrule
Uniform pooling (ours) & \textbf{93.65} & \textbf{55.98} & \textbf{54.66} & \textbf{57.98}\\
Learned pooling & 89.93 & 55.85 & 53.86 & 57.74\\
\midrule
No query spatial cues & 84.45 & 54.30 & 52.39 & 56.98\\
CLIP-only reader & 70.04 & 55.13 & 53.14 & 56.48\\
DINOv2-only reader & 89.57 & 55.94 & 54.65 & 57.80\\
No selector loss & 89.73 & 55.80 & 53.97 & 57.94\\
No change loss & 91.32 & 55.80 & 54.12 & 57.50\\
\bottomrule
\end{tabular*}
\end{table}

\noindent\textbf{Example pooling.}
We test whether example locations need different weights. Uniform pooling gives the highest mean mIoU on all four datasets, beating learned pooling by 3.72 points on EditCLEVR and 0.13--0.80 points on natural images. We therefore use uniform pooling.

\noindent\textbf{Spatial cues from the query pair.}
The decoder ablation measures how much spatial cues help localize the requested change. Without the query's DINOv2 spatial features and all-change probability map, mIoU drops by 5.48 points on EditCLEVR and 0.76--1.55 points on natural images.

\noindent\textbf{Visual features.}
To isolate each encoder's contribution, we test single-encoder readers. Relative to the two-encoder reader with learned pooling, CLIP alone costs 19.89 mIoU points on EditCLEVR, while DINOv2 alone costs only 0.36. DINOv2 alone also matches or slightly exceeds that reference on all three natural datasets.

\noindent\textbf{Auxiliary supervision.}
To test how direct supervision of the example selector and the all-change head affects localization, we remove one loss at a time. Mean mIoU changes little without the selector loss. Without the all-change loss, mean mIoU rises on EditCLEVR but falls on MaskEdit. We keep both losses to teach the selector where to read the example and the change head where any change occurs, without adding a step at inference.

\section{Conclusion}
In this paper, we propose VECD, a new framework that uses an unannotated image pair to specify the change to detect. The method first learns attribute states, then uses the demonstrated change to guide detection in another pair. Example-conditioned evidence and spatial cues from the query pair jointly determine the output mask. Experiments demonstrate selective detection among simultaneous changes and cross-dataset querying without text, example masks, or target-domain fine-tuning. A fixed example can guide many queries, while separate holdout tests show generalization to change families absent from paired training. Our work turns visual examples into reusable queries for finding the changes that users care about.
\label{maintextend}
\section*{LLM Usage Statement}
We used the OpenAI Codex model to assist with text writing and editing (e.g., language polishing, restructuring, and drafting prose) as well as code implementation. In these stages, the LLM was used to edit or refine author-produced content. We did not use AI tools for research ideation or other aspects of the study. The authors carefully reviewed and verified all AI-assisted content. All authors take full responsibility for all results and statements.
{\small
\bibliographystyle{ieeenat_fullname}
\bibliography{references}

\begin{thebibliography}{61}
\providecommand{\natexlab}[1]{#1}
\providecommand{\url}[1]{\texttt{#1}}
\expandafter\ifx\csname urlstyle\endcsname\relax
  \providecommand{\doi}[1]{doi: #1}\else
  \providecommand{\doi}{doi: \begingroup \urlstyle{rm}\Url}\fi

\bibitem[Alcantarilla et~al.(2016)Alcantarilla, Stent, Ros, Arroyo, and
  Gherardi]{alcantarilla2016}
Pablo~F. Alcantarilla, Simon Stent, German Ros, Roberto Arroyo, and Riccardo
  Gherardi.
\newblock Street-view change detection with deconvolutional networks.
\newblock In \emph{Robotics: Science and Systems}, 2016.

\bibitem[Ba et~al.(2016)Ba, Kiros, and Hinton]{layernorm2016}
Jimmy~Lei Ba, Jamie~Ryan Kiros, and Geoffrey~E. Hinton.
\newblock Layer normalization.
\newblock \emph{arXiv preprint arXiv:1607.06450}, 2016.

\bibitem[Bai et~al.(2024)Bai, Chow, Yang, Li, Li, Zhang, and
  Yan]{humanedit2024}
Jinbin Bai, Wei Chow, Ling Yang, Xiangtai Li, Juncheng Li, Hanwang Zhang, and
  Shuicheng Yan.
\newblock {HumanEdit}: A high-quality human-rewarded dataset for
  instruction-based image editing.
\newblock \emph{arXiv preprint arXiv:2412.04280}, 2024.

\bibitem[Bandara and Patel(2022)]{changeformer2022}
Wele Gedara~Chaminda Bandara and Vishal~M. Patel.
\newblock A transformer-based siamese network for change detection.
\newblock In \emph{IGARSS}, pages 207--210, 2022.

\bibitem[Bar et~al.(2022)Bar, Gandelsman, Darrell, Globerson, and
  Efros]{visualprompt2022}
Amir Bar, Yossi Gandelsman, Trevor Darrell, Amir Globerson, and Alexei~A.
  Efros.
\newblock Visual prompting via image inpainting.
\newblock In \emph{NeurIPS}, 2022.

\bibitem[Brooks et~al.(2023)Brooks, Holynski, and Efros]{instructpix2023}
Tim Brooks, Aleksander Holynski, and Alexei~A. Efros.
\newblock {InstructPix2Pix}: Learning to follow image editing instructions.
\newblock In \emph{CVPR}, pages 18392--18402, 2023.

\bibitem[Caron et~al.(2021)Caron, Touvron, Misra, J{\'e}gou, Mairal,
  Bojanowski, and Joulin]{dino2021}
Mathilde Caron, Hugo Touvron, Ishan Misra, Herv{\'e} J{\'e}gou, Julien Mairal,
  Piotr Bojanowski, and Armand Joulin.
\newblock Emerging properties in self-supervised vision transformers.
\newblock In \emph{ICCV}, pages 9650--9660, 2021.

\bibitem[Chen et~al.(2022)Chen, Qi, and Shi]{bitcd2022}
Hao Chen, Zipeng Qi, and Zhenwei Shi.
\newblock Remote sensing image change detection with transformers.
\newblock \emph{IEEE Transactions on Geoscience and Remote Sensing},
  60:\penalty0 1--14, 2022.

\bibitem[Cho et~al.(2024)Cho, Shin, Hong, Arnab, Seo, and Kim]{catseg2024}
Seokju Cho, Heeseong Shin, Sunghwan Hong, Anurag Arnab, Paul~Hongsuck Seo, and
  Seungryong Kim.
\newblock {CAT-Seg}: Cost aggregation for open-vocabulary semantic
  segmentation.
\newblock In \emph{CVPR}, pages 4113--4123, 2024.

\bibitem[Daudt et~al.(2018)Daudt, Le~Saux, and Boulch]{daudt2018}
Rodrigo~Caye Daudt, Bertrand Le~Saux, and Alexandre Boulch.
\newblock Fully convolutional siamese networks for change detection.
\newblock In \emph{ICIP}, pages 4063--4067, 2018.

\bibitem[Dosovitskiy et~al.(2021)Dosovitskiy, Beyer, Kolesnikov, Weissenborn,
  Zhai, Unterthiner, Dehghani, Minderer, Heigold, Gelly, Uszkoreit, and
  Houlsby]{vit2021}
Alexey Dosovitskiy, Lucas Beyer, Alexander Kolesnikov, Dirk Weissenborn,
  Xiaohua Zhai, Thomas Unterthiner, Mostafa Dehghani, Matthias Minderer, Georg
  Heigold, Sylvain Gelly, Jakob Uszkoreit, and Neil Houlsby.
\newblock An image is worth 16x16 words: Transformers for image recognition at
  scale.
\newblock In \emph{ICLR}, 2021.

\bibitem[Guo et~al.(2022)Guo, Wang, and Laaksonen]{clip4idc2022}
Zixin Guo, Tzu-Jui Wang, and Jorma Laaksonen.
\newblock {CLIP4IDC}: {CLIP} for image difference captioning.
\newblock In \emph{AACL-IJCNLP}, pages 33--42, 2022.

\bibitem[Hendrycks and Gimpel(2016)]{gelu2016}
Dan Hendrycks and Kevin Gimpel.
\newblock Gaussian error linear units ({GELUs}).
\newblock \emph{arXiv preprint arXiv:1606.08415}, 2016.

\bibitem[Hermann et~al.(2023)Hermann, Saha, and Zhu]{filtering2023}
Martin Hermann, Sudipan Saha, and Xiao~Xiang Zhu.
\newblock Filtering specialized change in a few-shot setting.
\newblock \emph{IEEE Journal of Selected Topics in Applied Earth Observations
  and Remote Sensing}, 16:\penalty0 1185--1196, 2023.

\bibitem[Jhamtani and Berg-Kirkpatrick(2018)]{spotdiff2018}
Harsh Jhamtani and Taylor Berg-Kirkpatrick.
\newblock Learning to describe differences between pairs of similar images.
\newblock In \emph{EMNLP}, pages 4024--4034, 2018.

\bibitem[Karnam and Sathish(2026)]{editclevr2026}
Anuraag~Gadehothur Karnam and Tarunesh Sathish.
\newblock {EditCLEVR}: A paired-scene intervention benchmark for compositional
  faithfulness of object-centric representations.
\newblock \emph{arXiv preprint arXiv:2607.22705}, 2026.

\bibitem[Kim et~al.(2021)Kim, Kim, Lee, Park, and Kim]{vacc2021}
Hoeseong Kim, Jongseok Kim, Hyungseok Lee, Hyunsung Park, and Gunhee Kim.
\newblock Viewpoint-agnostic change captioning with cycle consistency.
\newblock In \emph{ICCV}, pages 2095--2104, 2021.

\bibitem[Kim and Kim(2025)]{gescf2025}
Jae-Woo Kim and Ue-Hwan Kim.
\newblock Towards generalizable scene change detection.
\newblock In \emph{CVPR}, pages 24463--24473, 2025.

\bibitem[Kirillov et~al.(2023)Kirillov, Mintun, Ravi, Mao, Rolland, Gustafson,
  Xiao, Whitehead, Berg, Lo, Doll{\'a}r, and Girshick]{sam2023}
Alexander Kirillov, Eric Mintun, Nikhila Ravi, Hanzi Mao, Chloe Rolland, Laura
  Gustafson, Tete Xiao, Spencer Whitehead, Alexander~C. Berg, Wan-Yen Lo, Piotr
  Doll{\'a}r, and Ross Girshick.
\newblock Segment anything.
\newblock In \emph{ICCV}, pages 4015--4026, 2023.

\bibitem[Korkmaz et~al.(2026)Korkmaz, Paranjape, de~Melo, and Patel]{rcd2026}
Yilmaz Korkmaz, Jay~N. Paranjape, Celso~M. de Melo, and Vishal~M. Patel.
\newblock Referring change detection in remote sensing imagery.
\newblock In \emph{WACV}, 2026.

\bibitem[Lai et~al.(2025)Lai, Juefei-Xu, Liu, Dai, Mehta, Zhu, Huang, Rehg,
  Lee, Zhang, and Xiao]{instamanip2025}
Bolin Lai, Felix Juefei-Xu, Miao Liu, Xiaoliang Dai, Nikhil Mehta, Chenguang
  Zhu, Zeyi Huang, James~M. Rehg, Sangmin Lee, Ning Zhang, and Tong Xiao.
\newblock Unleashing in-context learning of autoregressive models for few-shot
  image manipulation.
\newblock In \emph{CVPR}, 2025.

\bibitem[Li et~al.(2022)Li, Weinberger, Belongie, Koltun, and Ranftl]{lseg2022}
Boyi Li, Kilian~Q. Weinberger, Serge Belongie, Vladlen Koltun, and Ren{\'e}
  Ranftl.
\newblock Language-driven semantic segmentation.
\newblock In \emph{ICLR}, 2022.

\bibitem[Liu et~al.(2020)Liu, Zhang, Zhang, and He]{ppnet2020}
Yongfei Liu, Xiangyi Zhang, Songyang Zhang, and Xuming He.
\newblock Part-aware prototype network for few-shot semantic segmentation.
\newblock In \emph{ECCV}, pages 142--158, 2020.

\bibitem[Liu et~al.(2024)Liu, Zhu, Li, Chen, Wang, and Shen]{matcher2024}
Yang Liu, Muzhi Zhu, Hengtao Li, Hao Chen, Xinlong Wang, and Chunhua Shen.
\newblock Matcher: Segment anything with one shot using all-purpose feature
  matching.
\newblock In \emph{ICLR}, 2024.

\bibitem[Loshchilov and Hutter(2019)]{adamw2019}
Ilya Loshchilov and Frank Hutter.
\newblock Decoupled weight decay regularization.
\newblock In \emph{ICLR}, 2019.

\bibitem[L{\"u}ddecke and Ecker(2022)]{clipseg2022}
Timo L{\"u}ddecke and Alexander Ecker.
\newblock Image segmentation using text and image prompts.
\newblock In \emph{CVPR}, pages 7086--7096, 2022.

\bibitem[Meng et~al.(2024)Meng, Lan, Li, Alvarez, Wu, and Jiang]{segic2024}
Lingchen Meng, Shiyi Lan, Hengduo Li, Jose~M. Alvarez, Zuxuan Wu, and Yu-Gang
  Jiang.
\newblock {SegIC}: Unleashing the emergent correspondence for in-context
  segmentation.
\newblock In \emph{ECCV}, pages 203--220, 2024.

\bibitem[Milletari et~al.(2016)Milletari, Navab, and Ahmadi]{vnet2016}
Fausto Milletari, Nassir Navab, and Seyed-Ahmad Ahmadi.
\newblock {V-Net}: Fully convolutional neural networks for volumetric medical
  image segmentation.
\newblock In \emph{3DV}, pages 565--571, 2016.

\bibitem[Min et~al.(2021)Min, Kang, and Cho]{hsnet2021}
Juhong Min, Dahyun Kang, and Minsu Cho.
\newblock Hypercorrelation squeeze for few-shot segmentation.
\newblock In \emph{ICCV}, pages 6941--6952, 2021.

\bibitem[Nguyen et~al.(2023)Nguyen, Li, Ojha, and Lee]{visii2023}
Thao Nguyen, Yuheng Li, Utkarsh Ojha, and Yong~Jae Lee.
\newblock Visual instruction inversion: Image editing via image prompting.
\newblock In \emph{NeurIPS}, 2023.

\bibitem[Nguyen et~al.(2024)Nguyen, Ojha, Li, Liu, and Lee]{editone2024}
Thao Nguyen, Utkarsh Ojha, Yuheng Li, Haotian Liu, and Yong~Jae Lee.
\newblock Edit one for all: Interactive batch image editing.
\newblock In \emph{CVPR}, pages 8271--8280, 2024.

\bibitem[Oquab et~al.(2024)Oquab, Darcet, Moutakanni, Vo, Szafraniec, Khalidov,
  Fernandez, Haziza, Massa, El-Nouby, et~al.]{dinov2}
Maxime Oquab, Timoth{\'e}e Darcet, Th{\'e}o Moutakanni, Huy Vo, Marc
  Szafraniec, Vasil Khalidov, Pierre Fernandez, Daniel Haziza, Francisco Massa,
  Alaaeldin El-Nouby, et~al.
\newblock {DINOv2}: Learning robust visual features without supervision.
\newblock \emph{Transactions on Machine Learning Research}, 2024.

\bibitem[Park et~al.(2019)Park, Darrell, and Rohrbach]{park2019}
Dong~Huk Park, Trevor Darrell, and Anna Rohrbach.
\newblock Robust change captioning.
\newblock In \emph{ICCV}, pages 4623--4632, 2019.

\bibitem[Park et~al.(2021)Park, Jang, Yoo, Lee, Kim, and Kim]{changesim2021}
Jin-Man Park, Jae-Hyuk Jang, Sahng-Min Yoo, Sun-Kyung Lee, Ue-Hwan Kim, and
  Jong-Hwan Kim.
\newblock {ChangeSim}: Towards end-to-end online scene change detection in
  industrial indoor environments.
\newblock In \emph{IROS}, 2021.

\bibitem[Qi et~al.(2026)Qi, Zhu, Li, Bi, Qi, Ma, and Yang]{dcsam2026}
Mengshi Qi, Pengfei Zhu, Xiangtai Li, Xiaoyang Bi, Lu Qi, Huadong Ma, and
  Ming-Hsuan Yang.
\newblock {DC-SAM}: In-context segment anything in images and videos via dual
  consistency.
\newblock \emph{IEEE Transactions on Pattern Analysis and Machine
  Intelligence}, 48\penalty0 (4):\penalty0 4642--4656, 2026.

\bibitem[Radford et~al.(2021)Radford, Kim, Hallacy, Ramesh, Goh, Agarwal,
  Sastry, Askell, Mishkin, Clark, Krueger, and Sutskever]{clip2021}
Alec Radford, Jong~Wook Kim, Chris Hallacy, Aditya Ramesh, Gabriel Goh,
  Sandhini Agarwal, Girish Sastry, Amanda Askell, Pamela Mishkin, Jack Clark,
  Gretchen Krueger, and Ilya Sutskever.
\newblock Learning transferable visual models from natural language
  supervision.
\newblock In \emph{ICML}, pages 8748--8763, 2021.

\bibitem[Radke et~al.(2005)Radke, Andra, Al-Kofahi, and Roysam]{radke2005}
Richard~J. Radke, Srinivas Andra, Omar Al-Kofahi, and Badrinath Roysam.
\newblock Image change detection algorithms: A systematic survey.
\newblock \emph{IEEE Transactions on Image Processing}, 14\penalty0
  (3):\penalty0 294--307, 2005.

\bibitem[Sachdeva and Zisserman(2023{\natexlab{a}})]{cyws2023}
Ragav Sachdeva and Andrew Zisserman.
\newblock The change you want to see.
\newblock In \emph{WACV}, 2023{\natexlab{a}}.

\bibitem[Sachdeva and Zisserman(2023{\natexlab{b}})]{cyws3d2023}
Ragav Sachdeva and Andrew Zisserman.
\newblock The change you want to see (now in 3d).
\newblock In \emph{ICCV Workshops}, pages 2052--2061, 2023{\natexlab{b}}.

\bibitem[Sakurada and Okatani(2015)]{sakurada2015}
Ken Sakurada and Takayuki Okatani.
\newblock Change detection from a street image pair using cnn features and
  superpixel segmentation.
\newblock In \emph{BMVC}, 2015.

\bibitem[Sakurada et~al.(2013)Sakurada, Okatani, and Deguchi]{sakurada2013}
Ken Sakurada, Takayuki Okatani, and Koichiro Deguchi.
\newblock Detecting changes in 3d structure of a scene from multi-view images
  captured by a vehicle-mounted camera.
\newblock In \emph{CVPR}, pages 137--144, 2013.

\bibitem[Sakurada et~al.(2018)Sakurada, Shibuya, and Wang]{sakurada2018}
Ken Sakurada, Mikiya Shibuya, and Weimin Wang.
\newblock Weakly supervised silhouette-based semantic scene change detection.
\newblock \emph{arXiv preprint arXiv:1811.11985}, 2018.

\bibitem[Shaban et~al.(2017)Shaban, Bansal, Liu, Essa, and Boots]{oslsm2017}
Amirreza Shaban, Shray Bansal, Zhen Liu, Irfan Essa, and Byron Boots.
\newblock One-shot learning for semantic segmentation.
\newblock In \emph{BMVC}, 2017.

\bibitem[Sheynin et~al.(2024)Sheynin, Polyak, Singer, Kirstain, Zohar, Ashual,
  Parikh, and Taigman]{emuedit2024}
Shelly Sheynin, Adam Polyak, Uriel Singer, Yuval Kirstain, Amit Zohar, Oron
  Ashual, Devi Parikh, and Yaniv Taigman.
\newblock {Emu Edit}: Precise image editing via recognition and generation
  tasks.
\newblock In \emph{CVPR}, pages 8871--8879, 2024.

\bibitem[Stauffer and Grimson(1999)]{stauffer1999}
Chris Stauffer and W.~E.~L. Grimson.
\newblock Adaptive background mixture models for real-time tracking.
\newblock In \emph{CVPR}, pages 246--252, 1999.

\bibitem[Sun et~al.(2023)Sun, Yang, Peng, Shen, Yang, Hu, Qiu, and
  Koike]{imagebrush2023}
Yasheng Sun, Yifan Yang, Houwen Peng, Yifei Shen, Yuqing Yang, Han Hu, Lili
  Qiu, and Hideki Koike.
\newblock {ImageBrush}: Learning visual in-context instructions for
  exemplar-based image manipulation.
\newblock In \emph{NeurIPS}, 2023.

\bibitem[Tian et~al.(2022)Tian, Zhao, Shu, Yang, Li, and Jia]{pfenet2022}
Zhuotao Tian, Hengshuang Zhao, Michelle Shu, Zhicheng Yang, Ruiyu Li, and Jiaya
  Jia.
\newblock Prior guided feature enrichment network for few-shot segmentation.
\newblock \emph{IEEE Transactions on Pattern Analysis and Machine
  Intelligence}, 44\penalty0 (2):\penalty0 1050--1065, 2022.

\bibitem[Varghese et~al.(2025)Varghese, Gao, and Hoskere]{viewdelta2025}
Subin Varghese, Joshua Gao, and Vedhus Hoskere.
\newblock {ViewDelta}: Scaling scene change detection through
  text-conditioning.
\newblock In \emph{ICCV Workshops}, 2025.

\bibitem[Vaswani et~al.(2017)Vaswani, Shazeer, Parmar, Uszkoreit, Jones, Gomez,
  Kaiser, and Polosukhin]{attention2017}
Ashish Vaswani, Noam Shazeer, Niki Parmar, Jakob Uszkoreit, Llion Jones,
  Aidan~N. Gomez, Lukasz Kaiser, and Illia Polosukhin.
\newblock Attention is all you need.
\newblock In \emph{NeurIPS}, 2017.

\bibitem[Wang et~al.(2019)Wang, Liew, Zou, Zhou, and Feng]{panet2019}
Kaixin Wang, Jun~Hao Liew, Yingtian Zou, Daquan Zhou, and Jiashi Feng.
\newblock {PANet}: Few-shot image semantic segmentation with prototype
  alignment.
\newblock In \emph{ICCV}, pages 9197--9206, 2019.

\bibitem[Wang et~al.(2025)Wang, Cveji{\'c}, Eldesokey, and Wonka]{editclip2025}
Qian Wang, Aleksandar Cveji{\'c}, Abdelrahman Eldesokey, and Peter Wonka.
\newblock {EditCLIP}: Representation learning for image editing.
\newblock In \emph{ICCV}, pages 15960--15970, 2025.

\bibitem[Wang et~al.(2023{\natexlab{a}})Wang, Wang, Cao, Shen, and
  Huang]{painter2023}
Xinlong Wang, Wen Wang, Yue Cao, Chunhua Shen, and Tiejun Huang.
\newblock Images speak in images: A generalist painter for in-context visual
  learning.
\newblock In \emph{CVPR}, pages 6830--6839, 2023{\natexlab{a}}.

\bibitem[Wang et~al.(2023{\natexlab{b}})Wang, Zhang, Cao, Wang, Shen, and
  Huang]{seggpt2023}
Xinlong Wang, Xiaosong Zhang, Yue Cao, Wen Wang, Chunhua Shen, and Tiejun
  Huang.
\newblock {SegGPT}: Towards segmenting everything in context.
\newblock In \emph{ICCV}, 2023{\natexlab{b}}.

\bibitem[Wang et~al.(2023{\natexlab{c}})Wang, Jiang, Lu, Shen, He, Chen, Wang,
  and Zhou]{promptdiffusion2023}
Zhendong Wang, Yifan Jiang, Yadong Lu, Yelong Shen, Pengcheng He, Weizhu Chen,
  Zhangyang Wang, and Mingyuan Zhou.
\newblock In-context learning unlocked for diffusion models.
\newblock In \emph{NeurIPS}, 2023{\natexlab{c}}.

\bibitem[Wu and He(2018)]{groupnorm2018}
Yuxin Wu and Kaiming He.
\newblock Group normalization.
\newblock In \emph{ECCV}, pages 3--19, 2018.

\bibitem[Xu et~al.(2026)Xu, Yong, Yang, Gong, and Lv]{maskflow2026}
Rui Xu, Yang Yong, Shunzi Yang, Ruihao Gong, and Chengtao Lv.
\newblock {MaskFlow}: Precise, consistent and seamless regional image editing.
\newblock \emph{arXiv preprint arXiv:2608.06929}, 2026.

\bibitem[Zhang et~al.(2019)Zhang, Lin, Liu, Yao, and Shen]{canet2019}
Chi Zhang, Guosheng Lin, Fayao Liu, Rui Yao, and Chunhua Shen.
\newblock {CANet}: Class-agnostic segmentation networks with iterative
  refinement and attentive few-shot learning.
\newblock In \emph{CVPR}, pages 5217--5226, 2019.

\bibitem[Zhang et~al.(2021)Zhang, Kang, Yang, and Wei]{cyctr2021}
Gengwei Zhang, Guoliang Kang, Yi Yang, and Yunchao Wei.
\newblock Few-shot segmentation via cycle-consistent transformer.
\newblock In \emph{NeurIPS}, 2021.

\bibitem[Zhang et~al.(2023)Zhang, Mo, Chen, Sun, and Su]{magicbrush2023}
Kai Zhang, Lingbo Mo, Wenhu Chen, Huan Sun, and Yu Su.
\newblock {MagicBrush}: A manually annotated dataset for instruction-guided
  image editing.
\newblock In \emph{NeurIPS Datasets and Benchmarks}, 2023.

\bibitem[Zhang et~al.(2024)Zhang, Jiang, Guo, Yan, Pan, Dong, Qiao, Gao, and
  Li]{persam2024}
Renrui Zhang, Zhengkai Jiang, Ziyu Guo, Shilin Yan, Junting Pan, Hao Dong, Yu
  Qiao, Peng Gao, and Hongsheng Li.
\newblock Personalize segment anything model with one shot.
\newblock In \emph{ICLR}, 2024.

\bibitem[Zhou et~al.(2022)Zhou, Loy, and Dai]{maskclip2022}
Chong Zhou, Chen~Change Loy, and Bo Dai.
\newblock Extract free dense labels from {CLIP}.
\newblock In \emph{ECCV}, 2022.

\end{thebibliography}
}
\onecolumn
\raggedbottom
\addtocontents{toc}{\protect\setcounter{tocdepth}{1}}
\setcounter{figure}{0}
\setcounter{table}{0}
\setcounter{equation}{0}
\renewcommand{\theHfigure}{supp.\arabic{figure}}
\renewcommand{\theHtable}{supp.\arabic{table}}
\renewcommand{\theHequation}{supp.\arabic{equation}}

\begin{center}
{\Large\bfseries\thetitle\par}\medskip
{\large Supplementary Material\par}
\end{center}
\appendix
\renewcommand{\thefigure}{S\arabic{figure}}
\renewcommand{\thetable}{S\arabic{table}}
\setcounter{tocdepth}{1}
\tableofcontents

\section{Datasets and attribute changes}
\label{app:datasets}
\begin{figure}[!ht]
\centering
\includegraphics[width=\linewidth]{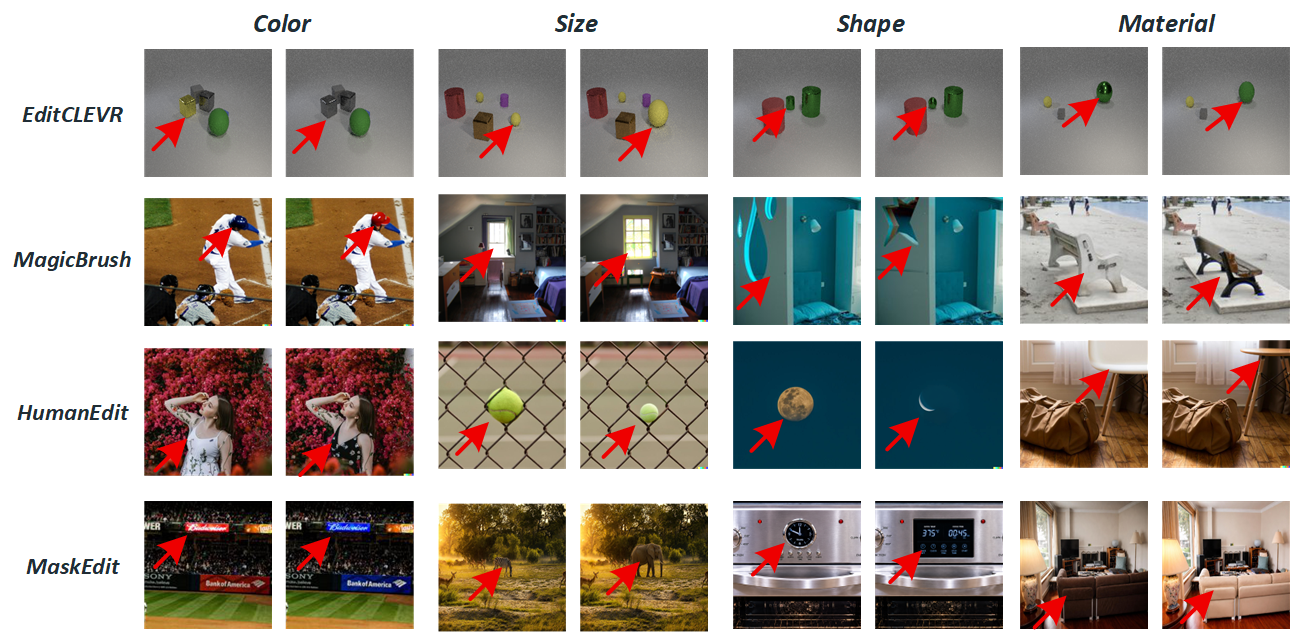}
\caption{\textbf{Attribute changes across datasets.} Each pair shows before and after images. Red arrows mark change regions and are not model inputs. In subsequent comparison figures, $M$: example mask; $A$: example-only adaptation; $T$: text; $U$: unconditional.}
\label{fig:supp_attributes}
\end{figure}

To illustrate the visual style and content of the four datasets, we select examples of color, size, shape, and material changes from each dataset, as shown in Figure~\ref{fig:supp_attributes}. MagicBrush, HumanEdit, and MaskEdit cover people, objects, and indoor and outdoor scenes, reflecting changes relevant to everyday visual applications. EditCLEVR adds rendered scenes, broadening the evaluation to a distinct visual domain. Together, these datasets evaluate the model's ability to match changes across diverse scenes and transfer from rendered images to natural images.

\section{How the five attribute queries are formed}
\label{app:queries}
This section details the query construction in the main paper's attribute-state reader. To compare the same property across different images, each query should specify \emph{which property to read}, while its output should reflect the local image content. We therefore use the five fixed descriptions in Table~\ref{tab:supp_descriptions}, rather than descriptions generated for individual images. Let $t_k$ denote the description for attribute $k$, where $k=1,\ldots,5$ indexes category, color, material, shape, and size. The frozen CLIP text encoder, including its pretrained text projection, maps each description to a normalized vector:
\begin{equation}
\bar u_k=\operatorname{norm}\!\left(f_{\rm text}(t_k)\right)
\in\mathbb{R}^{512}.
\label{eq:supp_fixed_queries}
\end{equation}
Here $f_{\rm text}$ denotes this frozen text-encoding function and $\operatorname{norm}$ denotes $\ell_2$ normalization. The vectors $\bar u_k$ are computed once and stored; for example, the color description remains the same for red and blue objects.
To adapt these fixed text features to the reader's visual memory, we apply a shared learnable projection and stack the projected vectors as rows:
\begin{equation}
u_k=W_U\bar u_k+b_U,\qquad
U=[u_1^\top;\ldots;u_5^\top]\in\mathbb{R}^{5\times512}.
\label{eq:supp_projected_queries}
\end{equation}
The matrix $W_U\in\mathbb{R}^{512\times512}$ and bias $b_U\in\mathbb{R}^{512}$ are the learned projection parameters. This mapping lets training adapt the property queries without changing CLIP or its stored text embeddings.
The resulting $U$ supplies the queries for the attribute-reading operation defined in the main paper.
During single-image training, $W_U$ and $b_U$ are updated with the reader, while the stored vectors $\bar u_k$ remain fixed. The entire reader is then frozen during paired training. The same $U$ is reused across all four images and all spatial locations, giving corresponding attribute slots a shared query. This construction supports comparison of the same properties across objects and scenes without requiring image-specific descriptions or text input at inference.

\begin{table}[!ht]
\centering\small
\begin{tabularx}{\linewidth}{@{}lX@{}}
\toprule
Reader slot & Exact description\\\midrule
Category & \texttt{the category of an object or a land cover region}\\
Color & \texttt{the color of an object}\\
Material & \texttt{the material of an object}\\
Shape & \texttt{the shape of an object}\\
Size & \texttt{the size of an object}\\
\bottomrule
\end{tabularx}
\caption{\textbf{Fixed descriptions used to initialize attribute queries.}}
\label{tab:supp_descriptions}
\end{table}

\section{Training losses and their supervision}
\label{app:training}
The main paper defines $\mathcal L_{\mathrm{static}}$ for learning attribute values and $\mathcal L_{\mathrm{pair}}$ for selecting changes. Here we thoroughly explain how the labels and masks in these objectives are used to compute each loss.

\paragraph{Attribute labels in $\mathcal L_{\mathrm{static}}$.}
The static loss \(\mathcal L_{\mathrm{static}}\) aims to make the reader's states represent specific properties rather than generic visual features. We supervise these states using EditCLEVR attribute labels. The query tells the reader \emph{what property to read}, while its output state must indicate \emph{which value is present}. We therefore use the region's ground-truth attribute value to supervise this output.
For color, we score the output state against fixed CLIP embeddings of red, blue, and the other candidate colors. These vectors are computed once with the frozen CLIP text encoder from descriptions. For attribute $k$, we collect these fixed value vectors as the rows of $B_k$. At a labeled location $p$, the normalized output state $Z_{p,k}$ gives the following score for candidate value $j$:
\begin{equation}
\ell_{p,k,j}=\frac{\langle Z_{p,k},b_{k,j}\rangle}{T},
\qquad T=0.07,
\end{equation}
Here $b_{k,j}$ is the $j$th normalized row of $B_k$.
Cross-entropy supervises the scores with ground-truth labels $y_k(p)$,
averaged over labeled locations for each attribute and then over
color, size, shape, and material. States are bilinearly upsampled
to $64\times64$ and renormalized; labels are resized with
nearest-neighbor interpolation. Background and unknown labels
($255$) are ignored.

These value-specific states are then used to compare the same attribute between the before and after images.

\paragraph{The teacher term in $\mathcal L_{\mathrm{static}}$.}
We use the teacher loss $\mathcal L_{\mathrm{teacher}}$ to preserve the reader's visual representation during attribute learning. Without this constraint, learning solely from EditCLEVR labels may cause the learned attribute states to become overly specific to rendered images, limiting their generalization to other image domains.

In detail, we follow the teacher setup described in the main paper. First, we copy the reader before single-image training and freeze this copy as the teacher. Both readers receive the same image. We then compare their normalized states at matching spatial locations and attribute slots:
\begin{equation}
\mathcal L_{\mathrm{teacher}}
=\frac{1}{5|\Omega|}
\sum_{p\in\Omega}\sum_{k=1}^{5}
\left(1-\langle Z_{p,k},Z^{\mathrm{teacher}}_{p,k}\rangle\right).
\end{equation}
Here $\Omega$ contains all grid locations across the training batch, and $Z^{\mathrm{teacher}}_{p,k}$ is the teacher state corresponding to $Z_{p,k}$. Both states are normalized, so their inner product is their cosine similarity. More similar states give a smaller loss. The comparison covers all five attributes, including category.

This constraint helps the reader learn attribute values while retaining visual information useful for change matching across objects and image domains.

\paragraph{Mask targets in $\mathcal L_{\mathrm{pair}}$.}
The paired loss $\mathcal L_{\mathrm{pair}}$ aims to select query changes that match the example and reject unrelated changes. Suppose the example shows a color change, while the query contains both a recolored object and an enlarged object. The target for $\mathcal L_{\mathrm{query}}$ marks only the recolored query object. The example mask used by $\mathcal L_{\mathrm{select}}$ marks the object demonstrating the color change in the example. In contrast, $\mathcal L_{\mathrm{change}}$ uses all-change masks: both changed query objects are foreground, and all changed example regions are foreground. If the query has no color change, its requested-change target is empty even if another property changes. These complementary targets help the detector learn both where changes occur and which changes match the example, supporting selection in scenes with multiple changes.

\paragraph{Computing the three mask losses.}
The main paper states that each mask loss combines binary cross-entropy and Dice loss. In the implementation, let $\widehat P$ be a predicted foreground probability map, $Y$ its binary target, and $\mathcal V$ the valid pixel set. We compute
\begin{equation}
\mathcal L_{\mathrm{mask}}(\widehat P,Y)
=\operatorname{BCE}_{\mathcal V}(\widehat P,Y)
+1-
\frac{2\sum_{p\in\mathcal V}\widehat P(p)Y(p)+1}
{\sum_{p\in\mathcal V}\widehat P(p)+\sum_{p\in\mathcal V}Y(p)+1}.
\end{equation}
The BCE term is averaged over valid pixels; the added $1$ smooths the Dice numerator and denominator. Ignored pixels contribute to neither term. The all-change loss averages the example and query mask losses. The three resulting terms are combined with the weights defined in the main paper, then averaged over training requests.

Example masks used as loss targets are training annotations, not extra test-time inputs. When a mask-guided training request already supplies the example mask to the model, we omit $\mathcal L_{\mathrm{select}}$ for that request. The reader stays frozen throughout paired training.

\section{Protocol-specific quantitative results}
\label{app:main_results}
To assess how consistently the main-paper results hold across training runs, we report means and sample standard deviations over three training seeds for each evaluation protocol. Reporting both captures average performance and training variability, helping distinguish stable improvements from results that depend on one run. 

Filtering's adaptation runs are averaged within each trained source model first. Matcher and GeSCF are training-free in this evaluation, so their scores are reported without training-seed standard deviations. Bold marks the largest displayed score for each dataset and metric within a protocol block, including ties; gray identifies VECD.

\paragraph{Same-dataset random examples.}
Table~\ref{tab:supp_same} shows that VECD's natural-image mIoU standard deviations are only $0.09$--$0.26$ points, while it retains the highest reported mIoU and F1 on all four datasets. Its EditCLEVR lead is much larger than the observed training variation. On HumanEdit, where the mIoU margin over ViewDelta is small, the F1 comparison provides additional evidence: $36.20\pm0.45$ versus $25.80\pm1.63$. These summaries support a repeatable gain.

\par\medskip\noindent
\begin{minipage}{\linewidth}
\captionsetup{type=table,hypcap=false}
\centering\footnotesize
\setlength{\tabcolsep}{2pt}
\begin{tabularx}{\linewidth}{@{}l*{8}{>{\raggedleft\arraybackslash}X}@{}}
\toprule
Method & \multicolumn{4}{c}{EditCLEVR} & \multicolumn{4}{c}{MagicBrush}\\
\cmidrule(lr){2-5} \cmidrule(lr){6-9}
 & mIoU & F1 & Re & Pre & mIoU & F1 & Re & Pre\\\midrule
Filtering$^{M,A}$ & \statcell{54.32}{\pm0.54} & \statcell{18.15}{\pm1.65} & \statcell{31.08}{\pm2.34} & \statcell{12.84}{\pm1.38} & \statcell{55.17}{\pm0.28} & \statcell{32.18}{\pm1.16} & \statcell{52.58}{\pm4.87} & \statcell{23.23}{\pm0.40}\\
Matcher$^M$ & \statcell{50.89}{} & \statcell{7.98}{} & \statcell{21.36}{} & \statcell{4.90}{} & \statcell{44.09}{} & \statcell{3.00}{} & \statcell{5.27}{} & \statcell{2.10}{}\\
SegIC$^M$ & \statcell{58.68}{\pm0.25} & \statcell{31.68}{\pm0.56} & \statcell{69.85}{\pm3.25} & \statcell{20.51}{\pm0.73} & \statcell{52.44}{\pm0.44} & \statcell{23.57}{\pm0.81} & \statcell{33.18}{\pm6.54} & \statcell{18.64}{\pm1.46}\\
DC-SAM$^M$ & \statcell{50.94}{\;(0.41)} & \statcell{9.27}{\;(0.68)} & \statcell{31.76}{\;(13.32)} & \statcell{5.43}{\;(0.18)} & \statcell{46.84}{\;(0.46)} & \statcell{6.41}{\;(1.44)} & \statcell{8.41}{\;(3.21)} & \statcell{5.18}{\;(0.70)}\\
ViewDelta$^T$ & \statcell{60.17}{\pm0.70} & \statcell{35.29}{\pm2.09} & \statcell{62.91}{\pm12.86} & \statcell{24.75}{\pm0.46} & \statcell{55.64}{\pm0.21} & \statcell{28.27}{\pm1.03} & \statcell{25.90}{\pm2.99} & \statcell{\textbf{31.46}}{\pm2.08}\\
GeSCF$^U$ & \statcell{57.97}{} & \statcell{29.03}{} & \statcell{44.37}{} & \statcell{21.57}{} & \statcell{54.23}{} & \statcell{32.85}{} & \statcell{68.07}{} & \statcell{21.65}{}\\
\midrule
\rowcolor{oursgray}
VECD (ours) & \statcell{\textbf{91.01}}{\pm1.19} & \statcell{\textbf{90.17}}{\pm1.42} & \statcell{\textbf{91.05}}{\pm1.31} & \statcell{\textbf{89.38}}{\pm3.23} & \statcell{\textbf{56.03}}{\pm0.09} & \statcell{\textbf{37.22}}{\pm0.24} & \statcell{\textbf{79.29}}{\pm2.40} & \statcell{24.32}{\pm0.13}\\
\bottomrule
\end{tabularx}
\par\medskip
\begin{tabularx}{\linewidth}{@{}l*{8}{>{\raggedleft\arraybackslash}X}@{}}
\toprule
Method & \multicolumn{4}{c}{HumanEdit} & \multicolumn{4}{c}{MaskEdit}\\
\cmidrule(lr){2-5} \cmidrule(lr){6-9}
 & mIoU & F1 & Re & Pre & mIoU & F1 & Re & Pre\\\midrule
Filtering$^{M,A}$ & \statcell{49.87}{\pm0.61} & \statcell{24.23}{\pm0.61} & \statcell{49.28}{\pm1.57} & \statcell{16.08}{\pm0.68} & \statcell{50.29}{\pm0.66} & \statcell{23.63}{\pm0.95} & \statcell{37.97}{\pm2.59} & \statcell{17.23}{\pm0.93}\\
Matcher$^M$ & \statcell{40.51}{} & \statcell{2.57}{} & \statcell{5.98}{} & \statcell{1.64}{} & \statcell{44.11}{} & \statcell{4.96}{} & \statcell{7.32}{} & \statcell{3.75}{}\\
SegIC$^M$ & \statcell{52.19}{\pm0.20} & \statcell{24.62}{\pm0.84} & \statcell{34.98}{\pm4.73} & \statcell{19.12}{\pm0.51} & \statcell{50.10}{\pm0.27} & \statcell{18.02}{\pm2.24} & \statcell{20.98}{\pm4.79} & \statcell{15.98}{\pm0.63}\\
DC-SAM$^M$ & \statcell{47.79}{\;(1.06)} & \statcell{11.04}{\;(0.87)} & \statcell{14.18}{\;(2.88)} & \statcell{9.03}{\;(2.31)} & \statcell{46.72}{\;(1.27)} & \statcell{12.11}{\;(1.55)} & \statcell{17.45}{\;(3.87)} & \statcell{9.27}{\;(2.25)}\\
ViewDelta$^T$ & \statcell{54.02}{\pm0.31} & \statcell{25.80}{\pm1.63} & \statcell{26.24}{\pm3.92} & \statcell{\textbf{25.67}}{\pm0.88} & \statcell{52.48}{\pm0.85} & \statcell{18.94}{\pm2.95} & \statcell{12.65}{\pm2.66} & \statcell{\textbf{38.65}}{\pm1.91}\\
GeSCF$^U$ & \statcell{53.38}{} & \statcell{32.03}{} & \statcell{63.28}{} & \statcell{21.44}{} & \statcell{54.75}{} & \statcell{27.09}{} & \statcell{22.24}{} & \statcell{34.65}{}\\
\midrule
\rowcolor{oursgray}
VECD (ours) & \statcell{\textbf{54.55}}{\pm0.16} & \statcell{\textbf{36.20}}{\pm0.45} & \statcell{\textbf{79.79}}{\pm2.15} & \statcell{23.41}{\pm0.23} & \statcell{\textbf{58.09}}{\pm0.26} & \statcell{\textbf{38.12}}{\pm0.71} & \statcell{\textbf{43.55}}{\pm1.92} & \statcell{33.92}{\pm0.71}\\
\bottomrule
\end{tabularx}
\caption{\textbf{Same-dataset random-example results (\%).}}
\label{tab:supp_same}
\end{minipage}
\par\medskip

\paragraph{Native multi-change queries.}
VECD reaches $92.20\pm0.89$ mIoU and $91.71\pm1.02$ F1 in Table~\ref{tab:supp_native}. Its low variation accompanies a large margin over the baselines, showing that the requested-change selection is not confined to one training run. The same 60 scenes yield 240 requests, including 84 absent-family requests. Targets contain only objects with the requested change.

\par\medskip\noindent
\begin{minipage}{\linewidth}
\captionsetup{type=table,hypcap=false}
\centering\small
\setlength{\tabcolsep}{3pt}
\begin{tabularx}{\columnwidth}{@{}l*{4}{>{\raggedleft\arraybackslash}X}@{}}
\toprule
Method & \multicolumn{4}{c}{EditCLEVR}\\
\cmidrule(lr){2-5}
 & mIoU & F1 & Re & Pre\\\midrule
Filtering$^{M,A}$ & \statcell{54.68}{\pm0.65} & \statcell{21.99}{\pm2.31} & \statcell{23.87}{\pm4.63} & \statcell{20.56}{\pm0.84}\\
Matcher$^M$ & \statcell{53.28}{} & \statcell{16.83}{} & \statcell{14.98}{} & \statcell{19.20}{}\\
SegIC$^M$ & \statcell{55.94}{\pm0.39} & \statcell{25.39}{\pm1.33} & \statcell{25.35}{\pm3.08} & \statcell{25.62}{\pm1.29}\\
DC-SAM$^M$ & \statcell{53.75}{\;(0.96)} & \statcell{19.68}{\;(4.14)} & \statcell{23.43}{\;(11.94)} & \statcell{16.96}{\;(0.73)}\\
ViewDelta$^T$ & \statcell{52.42}{\pm4.12} & \statcell{11.90}{\pm14.13} & \statcell{8.74}{\pm11.56} & \statcell{32.05}{\pm5.65}\\
\midrule
\rowcolor{oursgray}
VECD (ours) & \statcell{\textbf{92.20}}{\pm0.89} & \statcell{\textbf{91.71}}{\pm1.02} & \statcell{\textbf{91.78}}{\pm1.75} & \statcell{\textbf{91.74}}{\pm3.37}\\
\bottomrule
\end{tabularx}
\caption{\textbf{Native multi-change results (\%).}}
\label{tab:supp_native}
\end{minipage}
\par\medskip

\paragraph{Cross-dataset random examples.}
Across both example sources, VECD has the highest reported scores on all four metrics in Table~\ref{tab:supp_cross}, with mIoU standard deviations of $0.17$--$0.26$ points. Changing the source from MagicBrush to EditCLEVR changes its mean mIoU by only $0.09$ points on HumanEdit and $0.02$ on MaskEdit. The small variation complements the main-paper transfer results: performance remains stable across training seeds and across these two example sources. Each source evaluates 1,816 HumanEdit and 1,432 MaskEdit requests; no target-domain training or calibration is performed.

\par\medskip\noindent
\begin{minipage}{\linewidth}
\captionsetup{type=table,hypcap=false}
\centering\footnotesize
\setlength{\tabcolsep}{2pt}
\begin{tabularx}{\textwidth}{@{}l*{8}{>{\raggedleft\arraybackslash}X}@{}}
\toprule
Method & \multicolumn{4}{c}{HumanEdit} & \multicolumn{4}{c}{MaskEdit}\\
\cmidrule(lr){2-5} \cmidrule(lr){6-9}
 & mIoU & F1 & Re & Pre & mIoU & F1 & Re & Pre\\\midrule
\multicolumn{9}{l}{\textbf{MagicBrush examples}}\\\midrule
Filtering$^{M,A}$ & \statcell{52.92}{\pm0.54} & \statcell{28.73}{\pm1.28} & \statcell{48.40}{\pm3.42} & \statcell{20.45}{\pm0.93} & \statcell{52.22}{\pm0.51} & \statcell{27.30}{\pm1.17} & \statcell{40.95}{\pm3.26} & \statcell{20.51}{\pm0.95}\\
Matcher$^M$ & \statcell{40.58}{} & \statcell{3.20}{} & \statcell{7.56}{} & \statcell{2.03}{} & \statcell{43.52}{} & \statcell{2.62}{} & \statcell{3.76}{} & \statcell{2.01}{}\\
SegIC$^M$ & \statcell{52.32}{\pm0.44} & \statcell{24.87}{\pm0.21} & \statcell{34.96}{\pm4.02} & \statcell{19.44}{\pm1.07} & \statcell{51.49}{\pm0.34} & \statcell{20.94}{\pm0.77} & \statcell{22.46}{\pm3.90} & \statcell{20.11}{\pm2.13}\\
DC-SAM$^M$ & \statcell{46.89}{\;(1.34)} & \statcell{10.02}{\;(1.41)} & \statcell{14.28}{\;(2.97)} & \statcell{7.72}{\;(2.55)} & \statcell{48.33}{\;(0.63)} & \statcell{14.79}{\;(1.30)} & \statcell{19.07}{\;(4.41)} & \statcell{12.07}{\;(1.39)}\\
\midrule
\rowcolor{oursgray}
VECD (ours) & \statcell{\textbf{54.63}}{\pm0.17} & \statcell{\textbf{36.39}}{\pm0.45} & \statcell{\textbf{80.34}}{\pm2.13} & \statcell{\textbf{23.52}}{\pm0.24} & \statcell{\textbf{58.03}}{\pm0.25} & \statcell{\textbf{37.97}}{\pm0.70} & \statcell{\textbf{43.24}}{\pm1.93} & \statcell{\textbf{33.88}}{\pm0.62}\\
\midrule
\multicolumn{9}{l}{\textbf{EditCLEVR examples}}\\\midrule
Filtering$^{M,A}$ & \statcell{47.37}{\pm0.38} & \statcell{12.94}{\pm0.85} & \statcell{20.13}{\pm1.84} & \statcell{9.55}{\pm0.61} & \statcell{48.41}{\pm1.03} & \statcell{16.43}{\pm2.11} & \statcell{23.05}{\pm2.93} & \statcell{12.83}{\pm1.93}\\
Matcher$^M$ & \statcell{46.23}{} & \statcell{1.62}{} & \statcell{1.55}{} & \statcell{1.71}{} & \statcell{47.37}{} & \statcell{3.91}{} & \statcell{2.88}{} & \statcell{6.07}{}\\
SegIC$^M$ & \statcell{52.43}{\pm0.20} & \statcell{24.32}{\pm0.88} & \statcell{31.97}{\pm3.21} & \statcell{19.68}{\pm0.50} & \statcell{51.32}{\pm0.36} & \statcell{19.50}{\pm0.48} & \statcell{19.20}{\pm2.66} & \statcell{20.27}{\pm2.49}\\
DC-SAM$^M$ & \statcell{48.32}{\;(0.37)} & \statcell{3.53}{\;(1.42)} & \statcell{2.12}{\;(1.06)} & \statcell{10.59}{\;(3.85)} & \statcell{47.86}{\;(0.26)} & \statcell{2.94}{\;(1.72)} & \statcell{1.71}{\;(1.34)} & \statcell{10.55}{\;(2.39)}\\
\midrule
\rowcolor{oursgray}
VECD (ours) & \statcell{\textbf{54.72}}{\pm0.21} & \statcell{\textbf{36.36}}{\pm0.51} & \statcell{\textbf{79.02}}{\pm2.29} & \statcell{\textbf{23.61}}{\pm0.29} & \statcell{\textbf{58.01}}{\pm0.26} & \statcell{\textbf{37.67}}{\pm0.74} & \statcell{\textbf{41.59}}{\pm1.90} & \statcell{\textbf{34.45}}{\pm0.63}\\
\bottomrule
\end{tabularx}
\caption{\textbf{Cross-dataset random-example results (\%).}}
\label{tab:supp_cross}
\end{minipage}
\par\medskip

\paragraph{Fixed-example reuse.}
Table~\ref{tab:supp_fixed} adds variation estimates to the fixed-example comparison. On HumanEdit, VECD's mIoU varies less than Filtering's ($0.19$ versus $1.09$ points), while its F1 is higher and more stable ($36.45\pm0.46$ versus $29.27\pm3.06$). On MaskEdit, the mIoU lead exceeds five points with a $0.24$-point standard deviation. The small HumanEdit mIoU margin alone should not be read as conclusive. Eight fixed MagicBrush examples give 3,632 HumanEdit and 2,864 MaskEdit requests per run.

\par\medskip\noindent
\begin{minipage}{\linewidth}
\captionsetup{type=table,hypcap=false}
\centering\footnotesize
\setlength{\tabcolsep}{2pt}
\begin{tabularx}{\textwidth}{@{}l*{8}{>{\raggedleft\arraybackslash}X}@{}}
\toprule
Method & \multicolumn{4}{c}{HumanEdit} & \multicolumn{4}{c}{MaskEdit}\\
\cmidrule(lr){2-5} \cmidrule(lr){6-9}
 & mIoU & F1 & Re & Pre & mIoU & F1 & Re & Pre\\\midrule
Filtering$^{M,A}$ & \statcell{54.32}{\pm1.09} & \statcell{29.27}{\pm3.06} & \statcell{39.16}{\pm5.42} & \statcell{23.42}{\pm2.19} & \statcell{52.38}{\pm0.20} & \statcell{24.41}{\pm0.46} & \statcell{28.55}{\pm0.94} & \statcell{21.37}{\pm0.58}\\
Matcher$^M$ & \statcell{40.89}{} & \statcell{2.01}{} & \statcell{4.41}{} & \statcell{1.30}{} & \statcell{43.10}{} & \statcell{2.61}{} & \statcell{3.96}{} & \statcell{1.95}{}\\
SegIC$^M$ & \statcell{51.75}{\pm0.28} & \statcell{24.76}{\pm1.19} & \statcell{38.66}{\pm3.44} & \statcell{18.22}{\pm0.57} & \statcell{51.04}{\pm0.14} & \statcell{21.05}{\pm0.98} & \statcell{25.04}{\pm3.63} & \statcell{18.31}{\pm0.60}\\
DC-SAM$^M$ & \statcell{45.59}{\;(1.80)} & \statcell{8.00}{\;(1.77)} & \statcell{12.59}{\;(2.75)} & \statcell{5.86}{\;(2.82)} & \statcell{46.63}{\;(1.27)} & \statcell{12.02}{\;(1.94)} & \statcell{17.43}{\;(4.70)} & \statcell{9.17}{\;(2.19)}\\
\midrule
\rowcolor{oursgray}
VECD (ours) & \statcell{\textbf{54.66}}{\pm0.19} & \statcell{\textbf{36.45}}{\pm0.46} & \statcell{\textbf{80.47}}{\pm2.02} & \statcell{\textbf{23.56}}{\pm0.26} & \statcell{\textbf{57.98}}{\pm0.24} & \statcell{\textbf{37.80}}{\pm0.68} & \statcell{\textbf{42.84}}{\pm1.89} & \statcell{\textbf{33.86}}{\pm0.64}\\
\bottomrule
\end{tabularx}
\caption{\textbf{Fixed-example reuse results (\%).}}
\label{tab:supp_fixed}
\end{minipage}
\par\medskip

\section{Same-dataset queries and fixed-example reuse}
\label{app:qualitative}
Visual comparisons clarify the localization behavior behind the aggregate results. The same-dataset and fixed-example predictions across four datasets complement the three-seed statistics in Appendix~\ref{app:main_results} with examples of recovered regions, false positives, and reuse across domains. In Figure~\ref{fig:supp_same}, examples and queries differ in objects and spatial layout, while VECD selects the requested region without an example mask.

\begin{figure}[!htp]
\centering
{\small\bfseries (a) MagicBrush and EditCLEVR\par}\smallskip
\includegraphics[width=\linewidth]{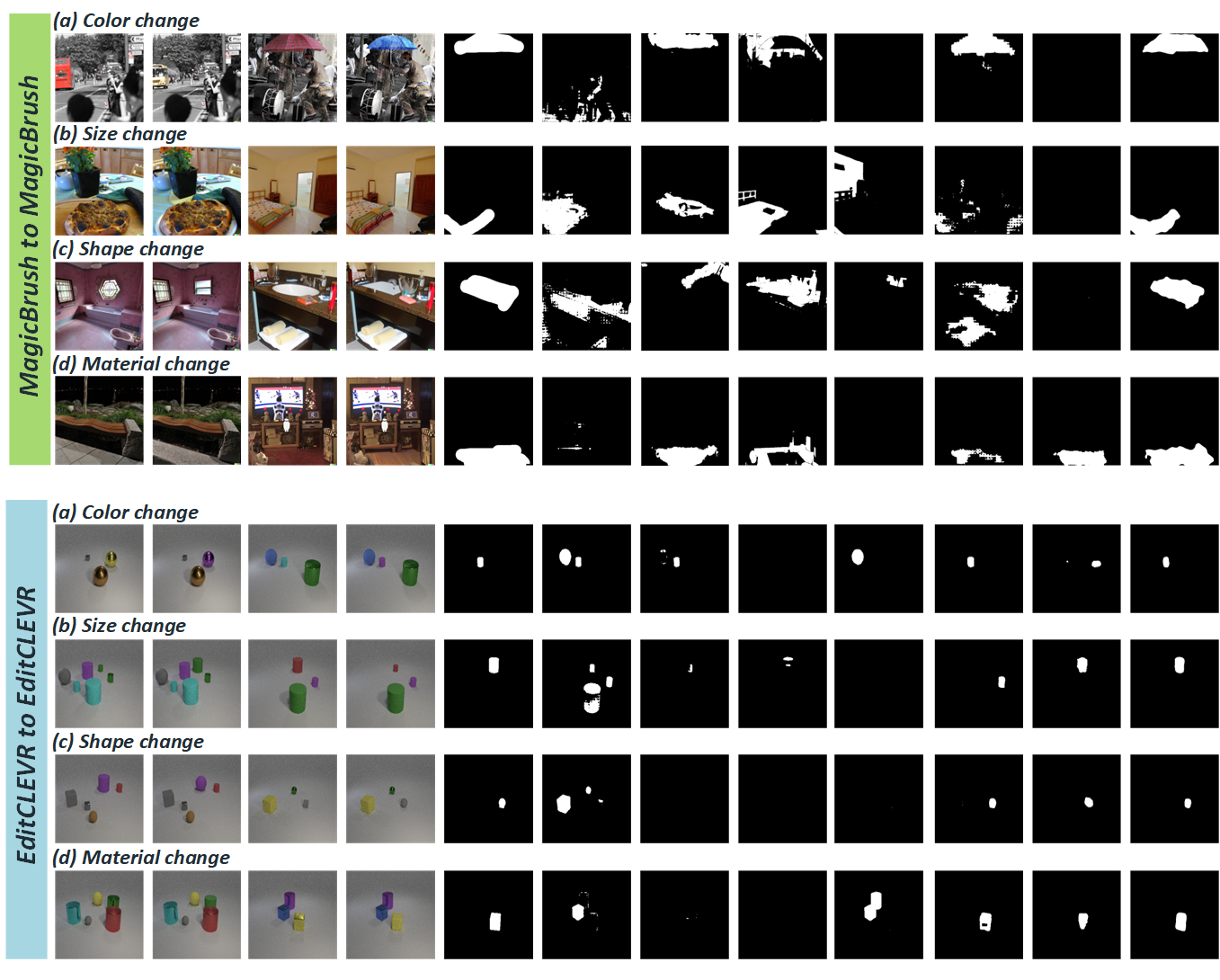}
\includegraphics[width=\linewidth]{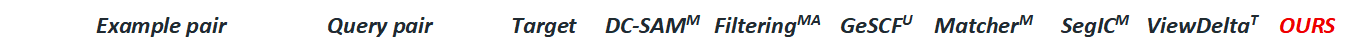}
\phantomcaption
\label{fig:supp_same}
\end{figure}

\begin{figure}[p]
\ContinuedFloat
\centering
{\small\bfseries (b) HumanEdit and MaskEdit\par}\smallskip
\includegraphics[width=\linewidth]{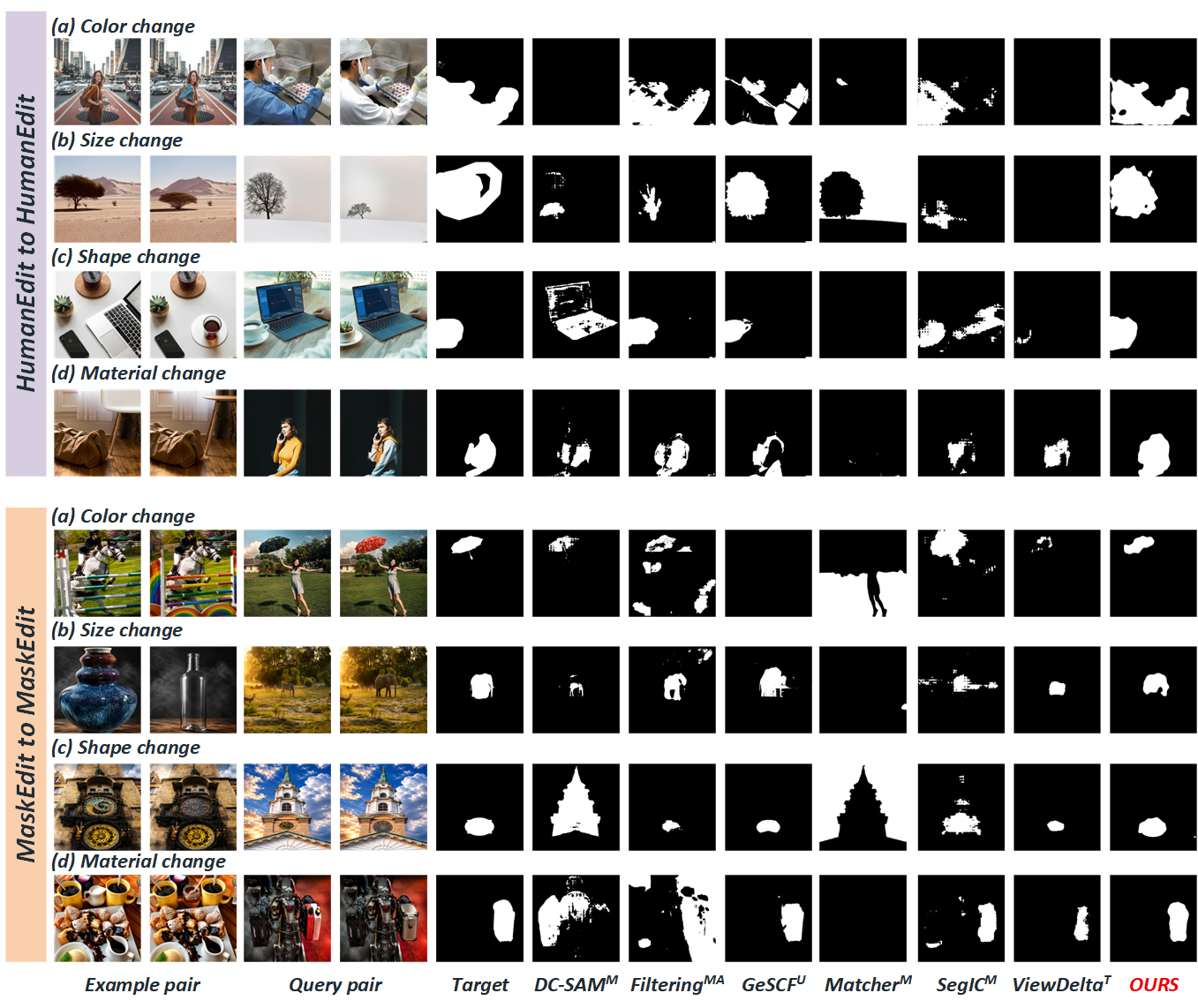}
\caption{\textbf{Same-dataset requests across four datasets.} Rows show color, size, shape, and material changes. Notation as in Figure~\ref{fig:supp_attributes}.}
\label{fig:supp_same_natural}
\end{figure}

\clearpage
\begingroup
\setlength{\intextsep}{6pt}
\captionsetup{skip=4pt}
\paragraph{A reusable natural-image request.}
Figure~\ref{fig:supp_fixed_natural} applies one MagicBrush color-change example to a rendered object, a bucket, clothing, and a skateboard. The objects, final colors, and target locations differ, so reusing the example requires matching the changed property rather than its appearance or position. These results highlight VECD's ability to reuse a single example for accurate change localization across diverse objects and image domains.

\begin{figure}[!htbp]
\centering
\includegraphics[width=\linewidth]{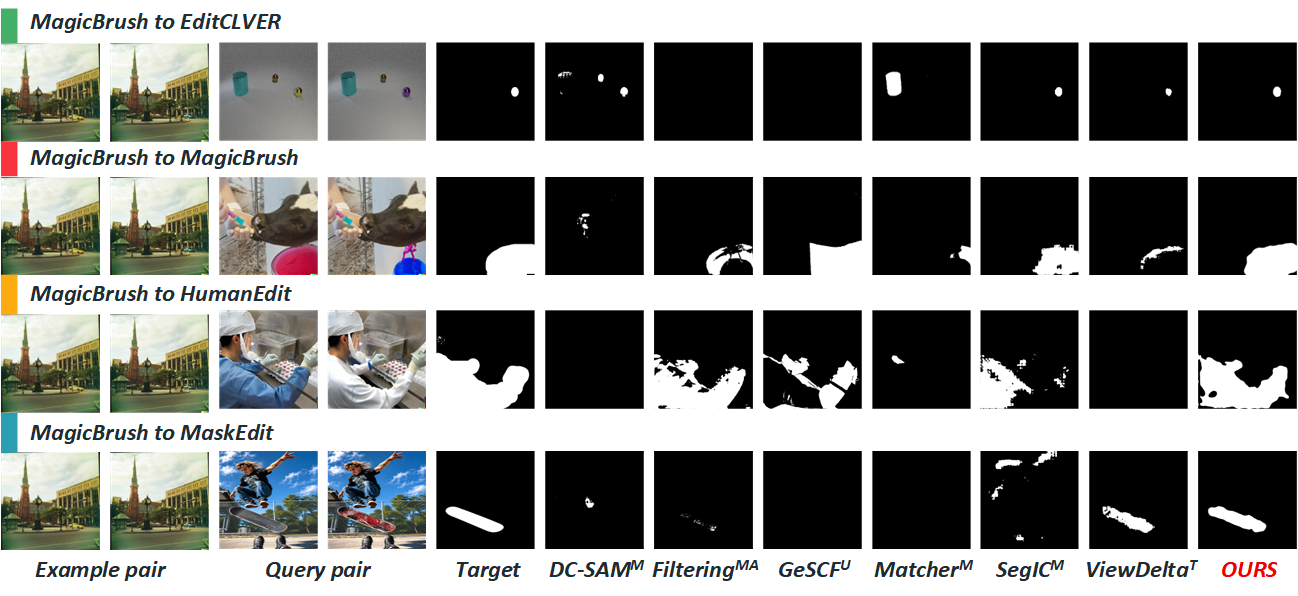}
\caption{\textbf{One fixed MagicBrush example queries four datasets.} Notation as in Figure~\ref{fig:supp_attributes}.}
\label{fig:supp_fixed_natural}
\end{figure}

\paragraph{A reusable rendered request.}
Figure~\ref{fig:supp_fixed_rendered} applies the same EditCLEVR size-change example to query pairs from all four datasets. This setting examines whether a request defined in a rendered scene can transfer to natural images with different objects, textures, and backgrounds. VECD localizes the resized rendered object as well as the edited window, food region, and control panel in natural images, demonstrating that one rendered example can guide change detection across visually distinct scenes.

\begin{figure}[!htbp]
\centering
\includegraphics[width=\linewidth]{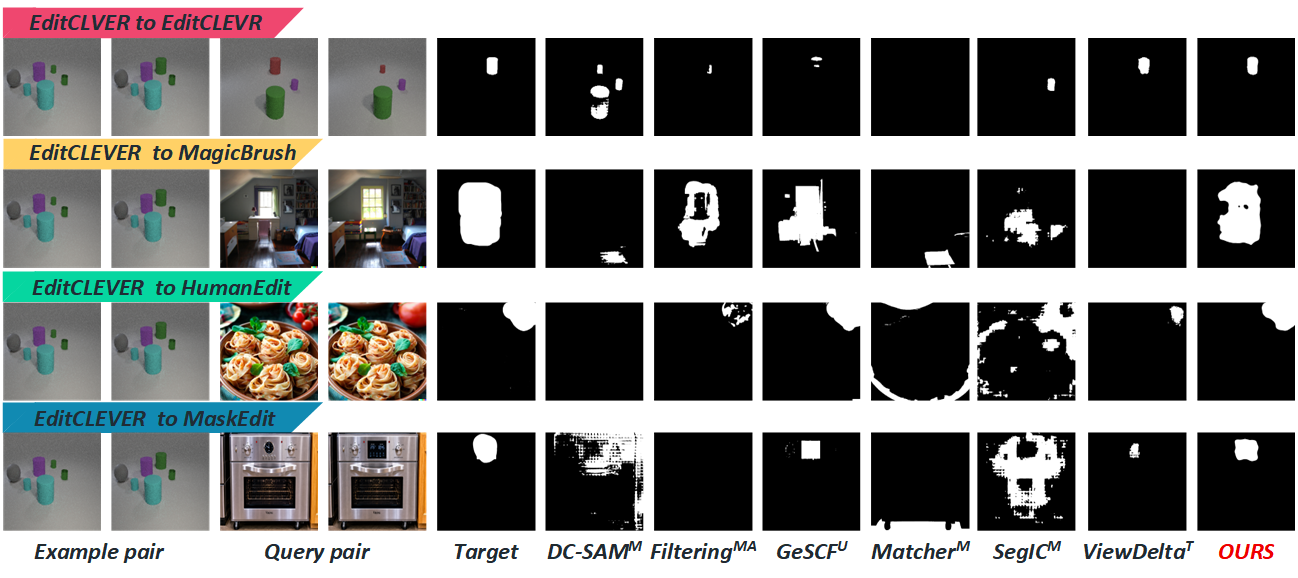}
\caption{\textbf{One fixed EditCLEVR example queries four datasets.} Notation as in Figure~\ref{fig:supp_attributes}.}
\label{fig:supp_fixed_rendered}
\end{figure}
\endgroup

\clearpage
\section{Ablation study}
\label{app:ablation}
Table~\ref{tab:supp_ablation} extends the main-paper ablations with each configuration's mean and sample standard deviation over three training seeds, allowing us to assess whether the findings are stable across training runs. Comparing both accuracy and consistency provides additional support for the final design choices. All configurations use the same evaluation examples. 

\begin{table}[!htbp]
\centering\footnotesize
\setlength{\tabcolsep}{2pt}
\begin{tabularx}{\linewidth}{@{}l*{8}{>{\raggedleft\arraybackslash}X}@{}}
\toprule
Variant & \multicolumn{4}{c}{EditCLEVR} & \multicolumn{4}{c}{MagicBrush}\\
\cmidrule(lr){2-5} \cmidrule(lr){6-9}
 & mIoU & F1 & Re & Pre & mIoU & F1 & Re & Pre\\\midrule
\rowcolor{oursgray}
Uniform pooling (ours) & \statcell{\textbf{93.65}}{\pm1.15} & \statcell{\textbf{93.25}}{\pm1.31} & \statcell{93.90}{\pm1.26} & \statcell{\textbf{92.65}}{\pm2.74} & \statcell{\textbf{55.98}}{\pm0.15} & \statcell{37.09}{\pm0.34} & \statcell{78.94}{\pm2.43} & \statcell{\textbf{24.24}}{\pm0.21}\\
\midrule
Learned pooling & \statcell{89.93}{\pm4.02} & \statcell{88.73}{\pm5.05} & \statcell{95.04}{\pm1.71} & \statcell{83.51}{\pm8.68} & \statcell{55.85}{\pm0.08} & \statcell{36.66}{\pm0.34} & \statcell{77.12}{\pm3.33} & \statcell{24.05}{\pm0.11}\\
No query spatial cues & \statcell{84.45}{\pm1.39} & \statcell{81.70}{\pm1.93} & \statcell{93.87}{\pm0.91} & \statcell{72.40}{\pm3.46} & \statcell{54.30}{\pm0.31} & \statcell{34.74}{\pm0.19} & \statcell{81.78}{\pm2.80} & \statcell{22.06}{\pm0.33}\\
CLIP-only reader & \statcell{70.04}{\pm1.85} & \statcell{57.78}{\pm3.63} & \statcell{85.14}{\pm5.70} & \statcell{44.10}{\pm5.61} & \statcell{55.13}{\pm0.26} & \statcell{33.99}{\pm0.20} & \statcell{65.50}{\pm4.64} & \statcell{22.98}{\pm0.40}\\
DINOv2-only reader & \statcell{89.57}{\pm2.07} & \statcell{88.40}{\pm2.55} & \statcell{94.40}{\pm1.00} & \statcell{83.19}{\pm4.43} & \statcell{55.94}{\pm0.11} & \statcell{36.58}{\pm1.26} & \statcell{75.34}{\pm9.96} & \statcell{24.22}{\pm0.19}\\
No selector loss & \statcell{89.73}{\pm2.46} & \statcell{88.58}{\pm3.08} & \statcell{\textbf{95.45}}{\pm0.35} & \statcell{82.72}{\pm5.08} & \statcell{55.80}{\pm0.10} & \statcell{\textbf{37.66}}{\pm0.06} & \statcell{\textbf{86.50}}{\pm1.54} & \statcell{24.07}{\pm0.11}\\
No change loss & \statcell{91.32}{\pm3.94} & \statcell{90.42}{\pm4.61} & \statcell{95.22}{\pm1.57} & \statcell{86.35}{\pm8.48} & \statcell{55.80}{\pm0.06} & \statcell{37.23}{\pm0.11} & \statcell{82.73}{\pm0.59} & \statcell{24.02}{\pm0.08}\\
\bottomrule
\end{tabularx}
\par\medskip
\begin{tabularx}{\linewidth}{@{}l*{8}{>{\raggedleft\arraybackslash}X}@{}}
\toprule
Variant & \multicolumn{4}{c}{HumanEdit} & \multicolumn{4}{c}{MaskEdit}\\
\cmidrule(lr){2-5} \cmidrule(lr){6-9}
 & mIoU & F1 & Re & Pre & mIoU & F1 & Re & Pre\\\midrule
\rowcolor{oursgray}
Uniform pooling (ours) & \statcell{\textbf{54.66}}{\pm0.19} & \statcell{\textbf{36.45}}{\pm0.46} & \statcell{80.47}{\pm2.02} & \statcell{23.56}{\pm0.26} & \statcell{\textbf{57.98}}{\pm0.24} & \statcell{37.80}{\pm0.68} & \statcell{42.84}{\pm1.89} & \statcell{\textbf{33.86}}{\pm0.64}\\
\midrule
Learned pooling & \statcell{53.86}{\pm0.76} & \statcell{35.05}{\pm0.96} & \statcell{78.88}{\pm1.43} & \statcell{22.54}{\pm0.91} & \statcell{57.74}{\pm0.36} & \statcell{37.21}{\pm0.86} & \statcell{41.93}{\pm1.31} & \statcell{33.45}{\pm0.77}\\
No query spatial cues & \statcell{52.39}{\pm0.67} & \statcell{33.22}{\pm0.58} & \statcell{81.35}{\pm2.88} & \statcell{20.89}{\pm0.65} & \statcell{56.98}{\pm0.21} & \statcell{37.94}{\pm0.06} & \statcell{\textbf{54.96}}{\pm2.74} & \statcell{29.00}{\pm0.71}\\
CLIP-only reader & \statcell{53.14}{\pm0.65} & \statcell{31.64}{\pm0.57} & \statcell{63.06}{\pm4.53} & \statcell{21.16}{\pm0.84} & \statcell{56.48}{\pm0.28} & \statcell{33.78}{\pm0.73} & \statcell{36.16}{\pm1.70} & \statcell{31.74}{\pm0.95}\\
DINOv2-only reader & \statcell{54.65}{\pm0.52} & \statcell{36.14}{\pm0.96} & \statcell{78.25}{\pm9.97} & \statcell{\textbf{23.57}}{\pm0.57} & \statcell{57.80}{\pm0.47} & \statcell{37.32}{\pm1.25} & \statcell{42.00}{\pm3.10} & \statcell{33.67}{\pm1.25}\\
No selector loss & \statcell{53.97}{\pm0.37} & \statcell{36.29}{\pm0.51} & \statcell{\textbf{87.45}}{\pm1.80} & \statcell{22.90}{\pm0.41} & \statcell{57.94}{\pm0.22} & \statcell{\textbf{38.33}}{\pm0.47} & \statcell{46.61}{\pm0.64} & \statcell{32.55}{\pm0.50}\\
No change loss & \statcell{54.12}{\pm0.35} & \statcell{35.86}{\pm0.65} & \statcell{82.21}{\pm0.93} & \statcell{22.93}{\pm0.46} & \statcell{57.50}{\pm0.22} & \statcell{36.75}{\pm0.52} & \statcell{41.85}{\pm0.70} & \statcell{32.76}{\pm0.50}\\
\bottomrule
\end{tabularx}
\caption{\textbf{Ablation results over three training seeds (\%).}}
\label{tab:supp_ablation}
\end{table}

\paragraph{Consistency of the final configuration.}
Uniform pooling retains the highest mean mIoU on all four datasets. Compared with learned pooling, its mIoU standard deviation falls from $4.02$ to $1.15$ points on EditCLEVR, from $0.76$ to $0.19$ on HumanEdit, and from $0.36$ to $0.24$ on MaskEdit. MagicBrush variation remains small for both choices. The added runs therefore support uniform pooling through both mean accuracy and improved consistency on three datasets.

\paragraph{What the variation adds.}
The final configuration also has more stable recall than the DINOv2-only variant on MagicBrush ($2.43$ versus $9.96$ points) and HumanEdit ($2.02$ versus $9.97$). Thus similar natural-image mIoU means can conceal different training sensitivity. Auxiliary-loss variants sometimes improve recall or F1, but none exceeds the final configuration's mean mIoU on any dataset. These results support our final configuration as a favorable balance between segmentation accuracy and consistency across training runs.
\end{document}